\documentclass[twoside]{article}

\usepackage[accepted]{aistats2021}
\usepackage[round]{natbib}
\renewcommand{\bibname}{References}
\renewcommand{\bibsection}{\subsubsection*{\bibname}}

\usepackage{hyperref}
\usepackage{url}

\usepackage{amsmath,amsfonts,bm,amssymb,mathtools,amsthm,bbm}
\usepackage{enumitem}
\usepackage{afterpage}

\usepackage{xcolor}
\newcommand{\vpara}[1]{\vspace{0.02in}\noindent\textbf{#1 }}

\newcommand*\phantomrel[1]{\mathrel{\phantom{#1}}}

\newcommand{\ind}{\mathbbm{1}}
\newcommand{\SN}{\lfloor N\rfloor}
\newcommand{\sbra}{\left(}
\newcommand{\sket}{\right)}
\newcommand{\reals}{\mathbb{R}}

\newtheorem{theorem}{Theorem}
\newtheorem{prop}{Proposition}

\newtheorem{lemma}{Lemma}
\newtheorem{definition}{Definition}

\def\eqref#1{equation~\ref{#1}}

\def\1{\bm{1}}

\def\vw{{\bm{w}}}

\DeclareMathAlphabet{\mathsfit}{\encodingdefault}{\sfdefault}{m}{sl}
\SetMathAlphabet{\mathsfit}{bold}{\encodingdefault}{\sfdefault}{bx}{n}

\usepackage{booktabs}

\begin{document}

\twocolumn[

\aistatstitle{Learning-to-Rank with Partitioned Preference: Fast Estimation for the Plackett-Luce Model}

\aistatsauthor{Jiaqi Ma\\jiaqima@umich.edu\\University of Michigan
    \And
    Xinyang Yi\\xinyang@google.com\\Google AI
    \And
    Weijing Tang\\weijtang@umich.edu\\University of Michigan
    \And
    Zhe Zhao\\zhezhao@google.com\\Google AI
    \AND
    Lichan Hong\\lichan@google.com\\Google AI
    \And
    Ed H. Chi\\edchi@google.com\\Google AI
    \And
    Qiaozhu Mei\\qmei@umich.edu\\University of Michigan}

\aistatsaddress{} ]

\begin{abstract}
We investigate the Plackett-Luce (PL) model based listwise learning-to-rank (LTR) on data with \textit{partitioned preference}, where a set of items are sliced into ordered and disjoint partitions, but the ranking of items within a partition is unknown. Given $N$ items with $M$ partitions, calculating the likelihood of data with partitioned preference under the PL model has a time complexity of $O(N+S!)$, where $S$ is the maximum size of the top $M-1$ partitions. This computational challenge restrains most existing PL-based listwise LTR methods to a special case of partitioned preference, \textit{top-$K$ ranking}, where the exact order of the top $K$ items is known. In this paper, we exploit a random utility model formulation of the PL model, and propose an efficient numerical integration approach for calculating the likelihood and its gradients with a time complexity $O(N+S^3)$. We demonstrate that the proposed method outperforms well-known LTR baselines and remains scalable through both simulation experiments and applications to real-world eXtreme Multi-Label classification tasks. 
\end{abstract}

\section{Introduction}
Ranking is a core problem in many information retrieval systems, such as recommender systems, search engines, and online advertising. The industry-scale ranking systems are typically applied to millions of items in a personalized way for billions of users. To meet the need of scalability and to exploit a huge amount of user feedback data, \textit{learning-to-rank} (LTR) has been the most popular paradigm for building the ranking system.
Existing LTR approaches can be categorized into three groups: \textit{pointwise}~\citep{gey1994inferring}, \textit{pairwise}~\citep{burges2005learning}, and \textit{listwise}~\citep{cao2007learning,taylor2008softrank} methods. The pointwise and pairwise LTR methods convert the ranking problem into regression or classification tasks on single or pairs of items respectively. As the real-world ranking data are often presented as (partially) ordered lists of items, the listwise LTR methods instead directly optimize objective functions defined on ranked lists of items, in order to preserve more information of the interrelations among items in a list.

One of the most well-known group of listwise LTR methods~\citep{cao2007learning,xia2008listwise} are based on the Plackett-Luce (PL) model~\citep{plackett1975analysis,luce1959individual}. These methods define their objective functions as the likelihood of the observed ranked list under a PL model. 
Despite being useful in many cases, a major limitation of such methods comes from the fact that, evaluating the likelihood of general partial rankings under a PL model is usually intractable for a large number of items. This computational challenge restricts the application of existing PL-based listwise LTR methods to limited special cases of partial rankings, such as \textit{top-$K$ ranking}, where the exact order of the top $K$ items is known. 

In this paper, we extend PL-based listwise LTR to a more general class of partial rankings, the \textit{partitioned preference}~\citep{lebanon2008non,lu2014effective}, defined as following: given $N$ items, partitioned preference slices the items into $M$ disjoint partitions, where order of items within each partition are unknown while the $M$ partitions have a global order. Partitioned preference not only is a strictly more general class of partial rankings compared to top-$K$ ranking, but also better characterizes real-world ranking data. For example, in a page of recommended items, we usually only observe binary clicks or a small number of ordinal ratings (e.g., 5-star rating) as user feedback but do not know the exact order among the clicked items or items with the same rating scale. However, computing the exact likelihood of data with partitioned preference under the PL model requires an intractable time complexity\footnote{The exact versions of time complexity measures mentioned in this section can be found in Section~\ref{sec:efficient}.} of $O(N + S!)$, where $S$ is the size of the largest partition among the top $M-1$ partitions. While there exist sampling-based methods~\citep{liu2019learning} that approximate the PL likelihood of partial rankings that are even more general than partitioned preferences, they cannot be directly adapted to the LTR setup where we usually need the gradients of the likelihood with respect to (w.r.t.) learnable parameters of a ranking model.

To overcome this computational challenge, we propose a novel numerical integration method. The key insight of our method is that, by exploiting a random utility model formulation of the PL model with Gumbel distribution~\citep{yellott1977relationship,mcfadden1978modeling}, we find that both the log-likelihood and its gradients can be re-written as the summation of multiple one-dimensional integrals. This finding enables the proposed numerical integration approach, which efficiently approximates the log-likelihood and the gradients. We formally demonstrate that, as the number of items grows, the overall time complexity of the proposed numerical approach is $O(N + \frac{1}{\epsilon}S^3)$ in order to maintain a constant level of numerical error $\epsilon$, which is much more efficient than the naive approach with the complexity $O(N+S!)$. We also discuss how our proposed approach might improve the generalized rank-breaking methods~\citep{khetan2018generalized}.

We evaluate the effectiveness of the proposed method through both simulation and experiments with real-world datasets. For simulation, we show that the proposed method can better recover the ground-truth parameters of a PL model compared to baseline methods, including a method~\citep{hino2010grouped} that approximates the PL likelihood with a tractable lower bound. We also test the proposed method on real-world extreme multilabel (XML) classification datasets~\citep{bhatia16extreme}. 
We show that the proposed method can efficiently train neural network ranking models for items at million-level, and outperforms other popular listwise and pairwise LTR baselines. 

\section{Related Work}
\subsection{Learning-to-Rank}
Our work falls in the area of LTR~\citep{liu2009learning}. The goal of LTR is to build machine learning models to rank a list of items for a given context (e.g., a user) based on the feature representation of the items and the context. The choice of the ranking objective plays an important role in learning the ranking models. Existing ranking objectives can be generally categorized in to three groups: pointwise~\citep{gey1994inferring}, pairwise~\citep{joachims2002optimizing,burges2005learning}, and listwise~\citep{cao2007learning,xia2008listwise,taylor2008softrank,christakopoulou2015collaborative,ai2018learning,wang2018lambdaloss,bruch2020stochastic}. The PL model has been widely used in listwise LTR methods~\citep{cao2007learning,xia2008listwise,schafer2018dyad}. 
However, to our best knowledge, existing PL-based listwise methods cannot be applied to partitioned preference data, due to the aforementioned computational complexity of evaluating the likelihood. Our work tackles the computational challenge with a novel numerical approach. Beyond the computational challenge, another major limitation of the PL-based listwise methods is that, the underlying independence of irrelevant alternatives (IIA) assumption of the PL model, is sometimes overly strong in real-world applications~\citep{seshadri2019fundamental,wilhelm2018practical,christakopoulou2015collaborative}. But more detailed discussions on the IIA assumption is out of the scope of this paper.

\vpara{XML classification as a ranking problem.}
Given features of each sample, the XML classification task requires a machine learning model to tag the most relevant subset of an extremely large label set. The XML classification tasks were initially established as a reformulation of ranking problems~\citep{agrawal2013multi,prabhu2014fastxml}, and the performance of which is primarily evaluated by various ranking metrics such as Precision@k or nDCG@k~\citep{bhatia16extreme}. The XML classification tasks are special cases of ranking with partitioned preference, where the class labels are considered items, and for each document its relevant labels form one partition and irrelevant labels form a second, lower-ranked partition. In this work, we apply the proposed method for ranking with partitioned preference to the XML classification datasets, and we find it achieving the state-of-the-art performance on datasets where the first partition, i.e., the set of relevant labels, is relatively large.

\subsection{Rank Aggregation}
\textit{Rank aggregation} aims to integrate multiple partial or full rankings into one ranking. The multiple rankings are considered as noisy samples from underlying ground truth ranking preferences. Rank aggregation is a broader research area that includes LTR as a subproblem. Statistical modeling is a popular approach for rank aggregation. Various statistical models~\citep{mallows1957non,luce1959individual,plackett1975analysis} are proposed to model the rank generation process in the real world. Among them, the PL model~\citep{luce1959individual,plackett1975analysis} is one of the most widely-used. Evaluating the likelihood of the PL model on various types of partial rankings has been widely studied~\citep{hunter2004mm,maystre2015fast,liu2019learning,yildiz2020fast,zhao2020learning}. However, we note that many of these studies~\citep{maystre2015fast,liu2019learning} are designed for ranking data without any features, and thus are not suitable for LTR tasks. The ones~\citep{yildiz2020fast,zhao2020learning} that can leverage sample features are not directly applicable to large-scale partitioned-preference data. It is worth noting that, our proposed method shares the motivation of approximating the intractable PL likelihood using sampling methods~\citep{liu2019learning}, as numerical integration is a special case of sampling. However, the integral form of the PL likelihood inspired by the connection between the PL model and Gumbel distribution makes our method more efficient than a general sampling method. 

\section{Approach}

\subsection{Problem Formulation: Learning PL Model from Partitioned Preference}
Suppose there are $N$ different items in total and we denote the set $\{1, \cdots, N\}$ by $\SN$. The PL model and the partitioned preference are formally defined below.
\begin{definition} [Plackett-Luce Model~\citep{plackett1975analysis,luce1959individual}]
Given the utility scores of the $N$ items, $\vw = [w_1, w_2, \cdots, w_N]^T$, the probability of observing a certain ordered list of these items, $(i_1, i_2, \cdots, i_N)$, is defined as
\begin{equation}
\label{eq:pl}
    p((i_1, i_2,\cdots, i_N); \vw) = \prod_{j=1}^N \frac{\exp(w_{i_j})}{\sum_{l=j}^N \exp(w_{i_l})}.
\end{equation}
\end{definition}

\begin{definition} [Partitioned Preference~\citep{lebanon2008non,lu2014effective}]
A group of $M$ disjoint partitions of $\SN$, $S_1, S_2,\cdots, S_M$, is called a partitioned preference if (a) $S_1 \succ \cdots \succ S_M$, where $S_m \succ S_{\tilde{m}}$ indicates that any item in $m$-th partition has a higher rank than items in the $\tilde{m}$-th partition; (b) the rank of items within the same partition is unknown.
\end{definition}
Clearly, $\cup_{m=1}^M S_m=\SN $ and $S_m \cap S_{\tilde{m}}=\emptyset$ for any $1\leq m\neq \tilde{m}\leq M$. We also denote the size of each partition $S_m$ as $n_m$, $m=1,\cdots, M$. Under a PL model parameterized by $\vw$ as defined in Eq.~(\ref{eq:pl}), the probability of observing such a partitioned preference $S_1 \succ  \cdots \succ S_M$ is given by
\begin{equation}
    \label{eq:pl-partition}
    \begin{split}
    & P(S_1 \succ  \cdots \succ S_M;\vw) \\
	= & \sum_{(i_1,\cdots,i_N)\in\Omega(S_1\succ\cdots\succ S_M;\SN)}\prod_{l=1}^{N}\frac{\exp(w_{i_l})}{\sum_{r=l}^N \exp(w_{i_r})},
	\end{split}
\end{equation}
where $\Omega(\cdot;\SN)$ is a function that maps a partial ranking to the set of all possible permutations of $\SN$ that are consistent with the given partial ranking.
 
Typically, the utility scores $\vw$ are themselves parameterized functions, e.g. neural networks, of the feature representation of the items and the context of ranking (e.g., a particular user). Suppose the features for item $i$ are denoted as $v_i$ and the item-independent context features are denoted as $x$. Then the utility score of $i$ for a given context (e.g., user) can be written as $w_i(x, v_i; \theta)$, where $\theta$ represents the neural network parameters\footnote{We simplify the notation $w_i(x, v_i;\theta)$ as $w_i$ for each $i\in\SN$ when there is no ambiguity.}. The problem of learning a PL model from data with partitioned preference can be formulated as maximizing the likelihood in Eq.~(\ref{eq:pl-partition}) over $\theta$. 

However, evaluating the likelihood function naively by Eq.~(\ref{eq:pl-partition}) requires a time complexity of $O(N+n_1!+\cdots+ n_{M-1}!)$, which is made clear in the form of Eq.~(\ref{eq:rewrite}) given by Lemma~\ref{lemma:rewrite}. This implies that the likelihood becomes intractable as long as one partition is mildly large (e.g., $20! > 10^{18}$). 

\begin{lemma}
    \label{lemma:rewrite}
    Let $\sigma(\cdot)$ be a function that maps a set of items to the set of all possible permutations of these items. Then Eq.~(\ref{eq:pl-partition}) can be re-written as
    \begin{equation}
        \label{eq:rewrite}
        \prod_{m=1}^{M-1} \left( \sum_{(i_1,\cdots, i_{n_m})\in \sigma(S_m)}\prod_{l=1}^{n_m}\frac{e^{w_{i_l}}}{\sum_{j\in R_m}e^{w_j}-\sum_{r=1}^{l-1}e^{w_{i_r}}}\right),
    \end{equation}
    which happens to be
    \begin{align*}
        \prod_{m=1}^{M-1} P(S_m\succ R_{m+1}),
    \end{align*}
    where $R_m$ is the set of items that do not belong to the top $m-1$ partitions, i.e. $R_m =\cup_{r=m}^{M}S_r$. 
\end{lemma}
 
\subsection{Efficient Evaluation of the Likelihood and Gradients}
\label{sec:efficient}
Next, we present how to efficiently evaluate the likelihood Eq.~(\ref{eq:likelihood}) and its gradients. We derive a numerical integral approach based on the random utility model formulation of the PL model with Gumbel distribution~\citep{yellott1977relationship,mcfadden1978modeling}. 

\paragraph{The random utility model formulation of PL.}
A random utility model assumes that, for each context, the utility of preferring the item $i\in\SN$ is a random variable $u_i = w_i+ \epsilon_i$. In particular, $w_i$ is the aforementioned parameterized utility function, and $\epsilon_i$ is a random noise term that contains all the unobserved factors affecting the individual's utility. When each $\epsilon_i$ independently follows a standard Gumbel distribution (or equivalently, each $u_i$ independently follows $\text{Gumbel}(w_i)$, a Gumbel distribution with the location parameter set to $w_i$), we have the following fact~\citep{yellott1977relationship}, for any permutation of items $(i_1, \cdots, i_{N})$, 
\[P(u_{i_1} > u_{i_2} > \cdots > u_{i_N}) = \prod_{j=1}^N\frac{\exp(w_{i_j})}{\sum_{k=j}^N \exp(w_{i_k})}.\]
It implies that, after sampling $N$ independent Gumbel variables, the ordered indices returned by sorting the Gumbel variables follow the PL model. Following this result, we have developed Proposition~\ref{prop:disjointab} that characterizes the preference for a given context between any two disjoint partitions.
\begin{prop}
\label{prop:disjointab}
Given a PL model parameterized by $\vw$, for any $A, B\subseteq \SN$ and $A\cap B =\emptyset$, the probability of $A\succ B$ is given by
\begin{align}
\label{eq:prop-disjointab}
P(A\succ B;\vw) &= P(\min_{a\in A}g_{w_a}>\max_{b\in B}g_{w_b}) \nonumber \\
&=\int_{u=0}^1\prod_{a\in A}(1-u^{\exp(w_a-w_{B})}) du,
\end{align}

where $g_w$ denotes a random variable following $\text{Gumbel}(w)$ and $w_B=\log \sum_{b\in B}\exp(w_b)$. 
\end{prop}
\citet{kool2020estimating} have shown a weaker version of Proposition~\ref{prop:disjointab} where $A\cup  B=\SN$. Here we extend it to the case where $A \cup B \subset \SN$, whose proof is given in Appendix~\ref{sec:sup-proof1}. 
In particular, the second equality in Eq.~(\ref{eq:prop-disjointab}) provides an efficient way of computing the likelihood of the preference for a given context between two disjoint partitions.

\paragraph{Compute the likelihood and gradients by numerical integral.}
Following the random utility model formulation of PL, we can compute the log-likelihood of the preference for a given context among $M$ partitions and its gradient efficiently by one-dimensional numerical integrals. For the likelihood function, it directly follows from Lemma~\ref{lemma:rewrite} and Proposition~\ref{prop:disjointab} that 
\begin{align}
\label{eq:likelihood}
\begin{split}
    &P (S_1 \succ  \cdots \succ S_M;\vw) \\
    =&\prod_{m=1}^{M-1}P(S_m \succ R_{m+1};\vw) \\
    =&\prod_{m=1}^{M-1}\int_{u=0}^1\prod_{i\in S_m}\sbra 1-u^{\exp\sbra w_i-w_{R_{m+1}}\sket}\sket du, 
\end{split}
\end{align}
where $w_{R_{m+1}}=\log \sum_{j\in R_{m+1}}\exp(w_j)$. Therefore, we can compute the log-likelihood function through $M-1$ one-dimensional numerical integration. From Lemma~\ref{lemma:gradient}, we can see that the gradients of the log-likelihood can also be obtained by $N-n_M$ numerical integrations using Eq.~(\ref{eq:gradient}).

\begin{lemma}
\label{lemma:gradient}
The gradients of the log-likelihood w.r.t. $\vw$ can be written as
\begin{align}
\label{eq:gradient}
\begin{split}
	&\nabla_{\vw}  \log P (S_1 \succ  \cdots \succ S_M ;\vw)\\ 
	=& - \sum_{m=1}^{M-1} \frac{1}{P(S_m \succ R_{m+1};\vw)} \sum_{i\in S_m} \nabla_{\vw}  \exp(w_i -w_{R_{m+1}}) \\
	& \phantomrel{==} \cdot  \int_{u=0}^1 \big[\prod_{j\in S_m} (1 - u^{ \exp(w_j -w_{R_{m+1}})})\big] \\
	& \phantomrel{=====} \cdot \frac{u^{ \exp(w_i -w_{R_{m+1}})} \log u}{1 - u^{ \exp(w_i -w_{R_{m+1}})}} du.
\end{split}
\end{align}
\end{lemma}

\vpara{Analysis of the computation cost.} Suppose the number of numerical integration intervals is set as $T$. Evaluating the likelihood~(\ref{eq:likelihood}) requires a time complexity $O(N + T(N-n_M))$ and evaluating the gradients~(\ref{eq:gradient}) requires a time complexity $O(N + T(\sum_{m=1}^{M-1}n_m^2))$. Next we quantify the minimum $T$ required by certain desired numerical error. The numerical error consists of the discretization error, which is caused by the discretization of the integral, and the round-off error, which is caused by the finite precision of the computer. While there is non-negligible round-off error if we calculate the numerical integration 
directly using formula (\ref{eq:likelihood}) and (\ref{eq:gradient}), this can be largely alleviated using common numerical tricks (see Appendix~\ref{sec:sup-num}). So we mainly focus on the analysis of the discretization error, which is given in Theorem~\ref{thm:error}. 
\begin{theorem}
\label{thm:error}
Assume that there exist some constants $c\in\reals$, and $C>1$ and $C = o(N^{1/6})$, such that, for any $1\le m\le M-1$,
\begin{equation}
    \label{eq:cond-c}
    2 < \exp(w_{R_m} +c) < C. 
\end{equation}
Then for any $\epsilon > 0$ and $m=1,\cdots, M-1$,
\begin{enumerate}[label=(\alph*)]
    \item we need at most $\frac{C^2(n_m+1)}{2\sqrt{3}\epsilon}$ intervals for the $m$-th integral in the likelihood~(\ref{eq:likelihood}) to have a discretization error smaller than $\epsilon$;
    \item if there exists some constant $0 < C_0 < 1$ and $\frac{1}{C_0}=o(\sqrt{N})$, such that, for each $i\in S_m$, defining $a=\exp(w_i + w_{R_{m+1}} + 2c)$ and $b = \exp(w_i + c)$, the following conditions hold,
    \begin{equation}
        \label{eq:condition}
        a > 4, a + 2b > 5, \text{ and } b > C_0,
    \end{equation}
    then we need at most $\frac{\sqrt{6} C^{11/2} n_m^2}{C_0^2\epsilon}$ intervals for the $(m, i)$-th integral in gradients~(\ref{eq:gradient}) to have a discretization error smaller than $\epsilon/n_m$.
\end{enumerate}
\end{theorem}
We note that the assumptions (\ref{eq:cond-c}) and (\ref{eq:condition}) in Theorem~\ref{thm:error} respectively require the largest and the smallest neural network output logit, plus a constant $c$, to be not too far away from zero\footnote{In practice, we find ranging from -10 to 10 is close enough to give good empirical results.}. These assumptions are easy to satisfy by first controlling the scale of the logits $\vw$, and then choosing a proper $c$ to center the logits ($c$ could be either positive or negative). We provide the proof of Theorem~\ref{thm:error} in Appendix~\ref{sec:thm1-proof}. 

Theorem~\ref{thm:error} implies that the computation cost of the proposed method is overall $O(N + \frac{1}{\epsilon}\sum_{m=1}^{M-1}n_m^3)$ to maintain a discretization error at most $\epsilon$ for both the likelihood and its gradients, which is clearly much more efficient than the naive approach with factorial terms. We further highlight several computational advantages of the proposed numerical approach. 1) The whole computation is highly parallelizable: the computation of the $T$ integrands and the product over $S_m$ within each integrand can all be done in parallel. 2) The number of intervals $T$ can be adjusted to control the trade-off between computation cost and accuracy\footnote{In our experiments in the main paper, we fix the hyper-parameters $T=10000$ and $c=5$ in all settings. We also provide}. 3) In large-scale ranking data, we often have $N \simeq n_M \gg \sum_{m=1}^{M-1}n_m$, thus $\sum_{m=1}^{M-1}n_m^3$ will be negligible for large $N$, resulting a linear complexity w.r.t. $N$. 

\subsection{Improving the Computational Efficiency of Generalized Rank-Breaking Methods}
We discuss the potential application of the proposed numerical approach to \textit{generalized rank-breaking methods}~\citep{khetan2018generalized} as a final remark of this section. 

While partitioned preference is a general class of partial rankings for a set of items, it is not able to represent the class of all possible partial rankings, which is also known as \textit{arbitrary pairwise preference}~\citep{lu2014effective,liu2019learning}. It is challenging to learn arbitrary pairwise preference using a listwise method. To the best of our knowledge, there is no scalable listwise method that is able to learn industry-scale PL-based ranking models. Pointwise and pairwise methods are able to deal with any types of partial rankings at the expense of lower statistical efficiency. Generalized rank-breaking methods~\citep{khetan2018generalized} are recently proposed to better trade-off the computational and statistical efficiency in LTR. 

An arbitrary pairwise preference of $N$ items can be represented as a directed acyclic graph (DAG) of $N$ nodes, where each node is an item and each directed edge represents the preference over a pair of items. The generalized rank-breaking methods first apply a graph algorithm to extract a \textit{maximal ordered partition} of $\SN$, $S_1 \succ S_2, \cdots, S_M$: a group of $M$ disjoint partitions of $\SN$ with largest possible M, such that the item preference in the M partitions is consistent with that of the DAG. One difference between data with partitioned preference and data with arbitrary pairwise preference is that the maximal ordered partition is not unique for the latter, as the maximal ordered partition does not preserve all relationships in the DAG. With the extracted partitions, we can maximize the likelihood of these partitions under a PL model to learn the model parameters.  \citet{khetan2018generalized} propose to calculate the likelihood as shown in Eq.~(\ref{eq:pl-partition}), which has a time complexity involves factorials of the partition sizes. To overcome this challenge for learning large-scale data, existing methods need to approximate the likelihood by dropping the top $M-1$ partitions with large sizes. In contrast, the proposed numerical approach in this paper can be directly applied to the likelihood evaluation step of generalized rank-breaking methods to significantly improve the computational efficiency.
\section{Experiments}

\begin{figure*}
	\centering
	\includegraphics[width=0.4\linewidth]{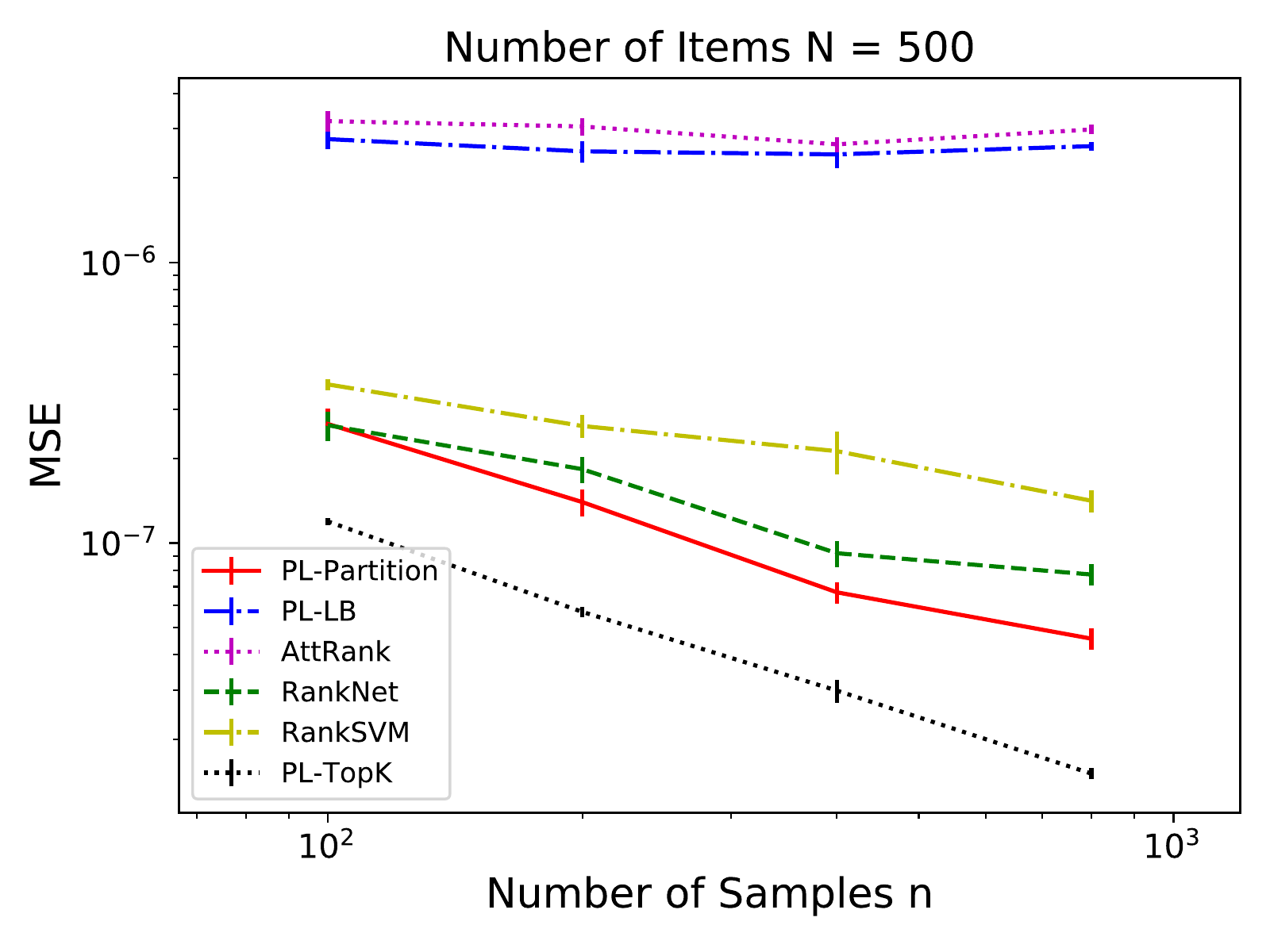}
	\includegraphics[width=0.4\linewidth]{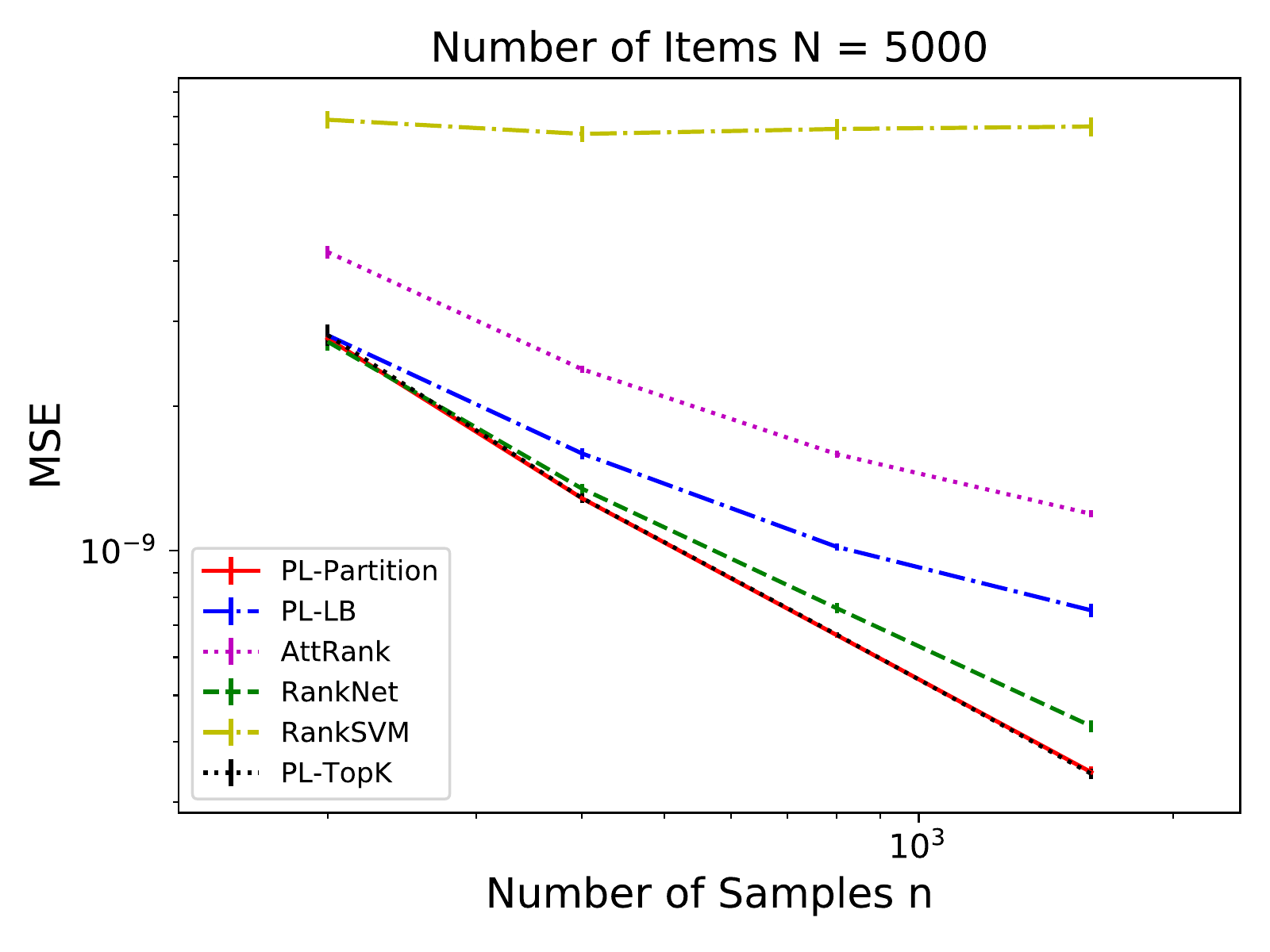}
	\caption{MSE of the estimated PL utility scores vs various numbers of items $N$ and number of samples $n$. Both x-axis and y-axis are in the logarithmic scale with base 10. The results are averaged over 5 different random seeds and error bars indicate the standard error of the mean.}
	\label{fig:sim-mse}
	\vskip -7pt
\end{figure*}

In this section, we report empirical results on both synthetic and real-world datasets. We compare the proposed method, denoted as \textbf{PL-Partition}, with two groups of baseline methods that can be applied to large-scale partitioned preference data. 

First, we consider two softmax-based listwise methods: \textbf{PL-LB}~\citep{hino2010grouped} and \textbf{AttRank}~\citep{ai2018learning}. PL-LB optimizes a lower bound of the likelihood of partitioned preference under the PL model. In particular, for each $m=1,\cdots,M-1$, the term $P(S_m\succ R_{m+1};\vw)$ in Eq.~(\ref{eq:likelihood}) is replaced by its lower bound $n_m!\prod_{i\in S_m}\left( \exp(w_i)/\sum_{j\in S_m\cup R_{m+1}}\exp(w_j)\right)$. AttRank optimizes the cross-entropy between the softmax outputs and an empirical probability based on the item relevance given by training labels. 

For the second group of baselines, we consider two popular pairwise methods: \textbf{RankNet}~\citep{burges2005learning} and \textbf{RankSVM}~\citep{joachims2002optimizing}. RankNet optimizes a logistic surrogate loss on each pair of items that can be compared. RankSVM optimizes a hinge surrogate loss on each pair of items that can be compared.

\subsection{Simulation}
\label{sec:sim}
We conduct experiments on synthetic data which is generated from a PL model. The goal of this simulation study is two-fold: 1) we investigate how accurate the proposed method can recover the ground truth utility scores of a PL model; 2) we empirically compare the computation costs of different methods over data with different scales. 

\vpara{Synthetic data generated from a PL model.}
We first generate a categorical probability simplex $p \in \Delta^{N-1}$ as the ground truth utility scores for $N$ items following $p = \text{softmax}(q)$ and $q_i \stackrel{\text{i.i.d.}}{\sim} \text{uniform}(0, \log N), 1\le i\le N$. Then we draw $n$ samples of full ranking from a PL model parameterized by $p$. Finally, we randomly split the full rankings into $M$ partitions and remove the order within each partition to get the partitioned preference. We note that the synthetic data is stateless (i.e., there is no feature for each sample), as this simulation focuses on the estimation of the PL utility scores $p$ rather than the relationship between $p$ and sample features. Further, in large-scale real-world applications, we often can only observe the order of limited items per sample. For example, a user can only consume a limited number of recommended items. To respect this pattern, we restrict the total number of items in top $M - 1$ partitions to be at most 500 regardless of $N$. We fix $M=4$ and generate data with varying $(N, n)$ and random seeds. 

\vpara{Experiment setups.} As the synthetic data is stateless, we only need to train $N$ free parameters, with each parameter corresponding to an item. We use the proposed method and baseline methods to respectively train the parameters using stochastic gradient descent with early-stopping. We use AdaGrad optimizer with initial learning rate of 0.1 for all methods. We report the mean squared error (MSE) between the softmax of these free parameters and the PL utility scores $p$ as a measure of how accurately different methods recover $p$. 

\vpara{MSE of the estimated PL utility scores.} Figure~\ref{fig:sim-mse} shows the MSE of the estimated PL utility scores by different methods over various $N$ and $n$. To better compare results across data with different numbers of items $N$, we further include an oracle reference method \textbf{PL-TopK}, which has access to the full ranking of the items in the top $M-1$ partitions and optimizes the corresponding PL likelihood.  First, as expected, the proposed PL-Partition method best recovers the ground truth utility scores in terms of MSE on all data configurations, as it numerically approximates the PL likelihood of the partitioned preference data. However, it is worth noting that PL-LB, while also trying to approximate the PL likelihood with a lower bound, performs even worse than pairwise methods when $N$ is small, which indicates the existing lower bound method is not sufficient to take the full advantage of the PL model.

\begin{figure*}
	\centering
	\includegraphics[width=0.4\linewidth]{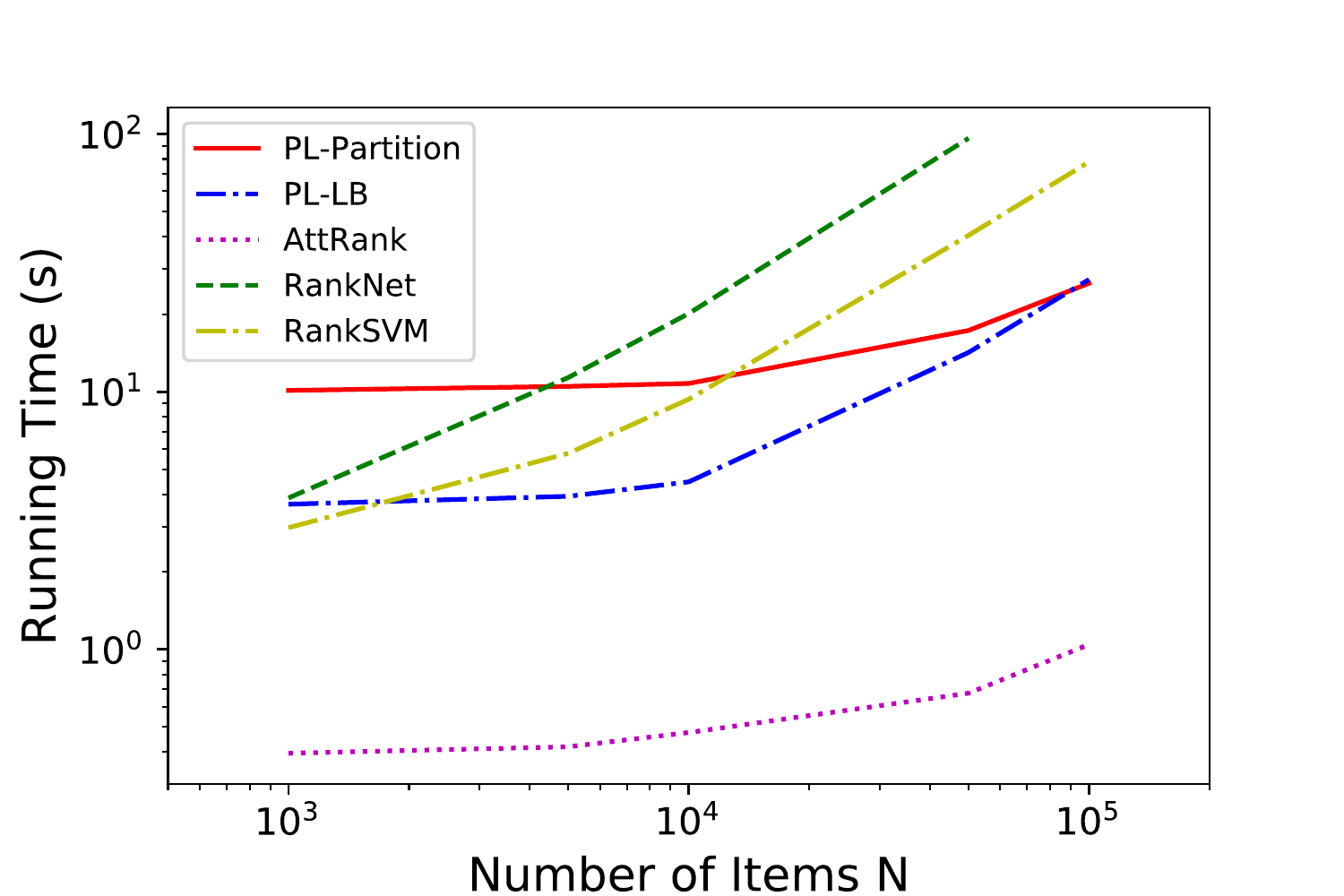}
	\includegraphics[width=0.4\linewidth]{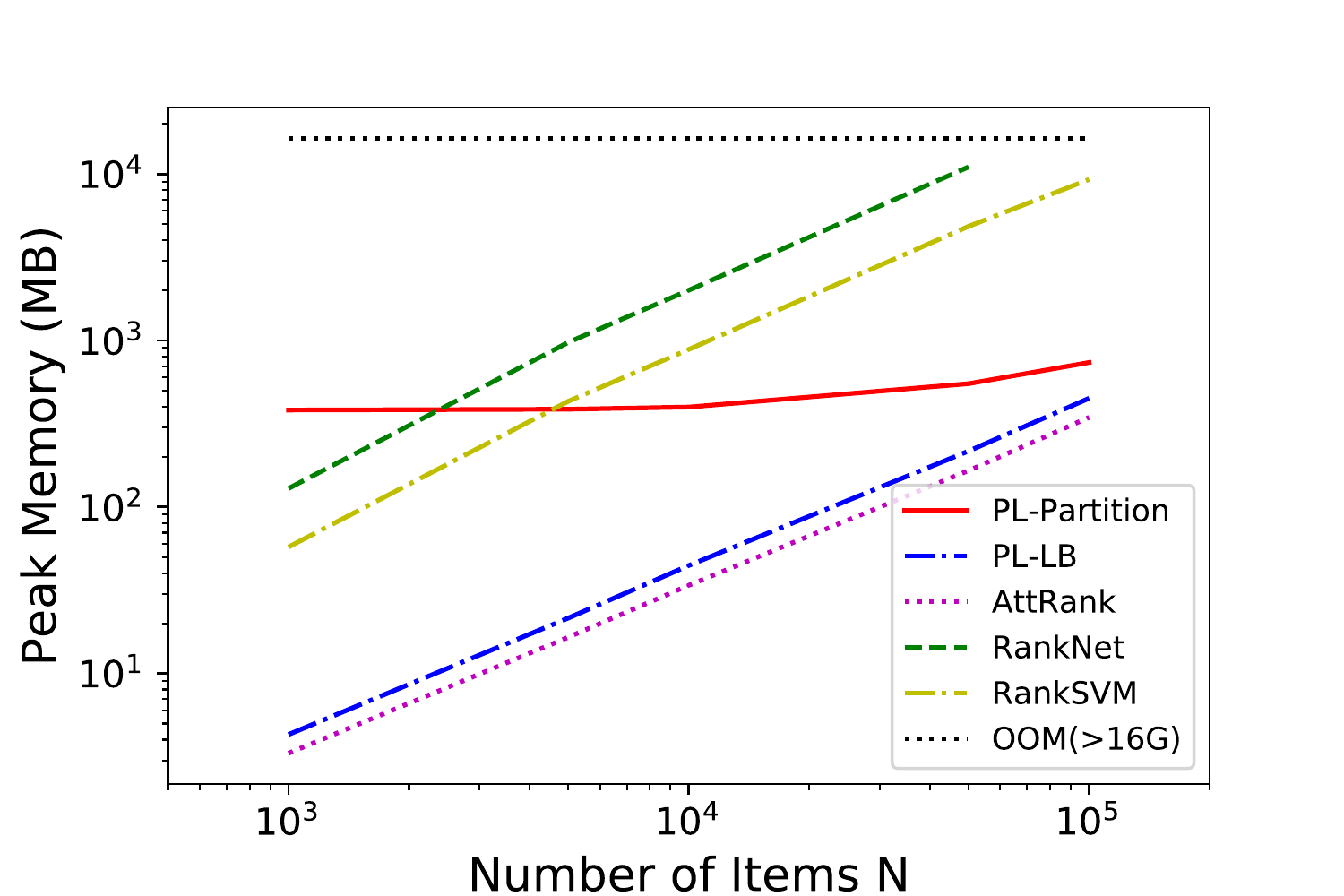}
	\caption{Time and memory cost of different methods for various number of items $N$ (average of 5 different random seeds, each running 1000 steps of stochastic gradient descent with batch size 20). Both x-axis and y-axis are in the logarithmic scale with base 10. The standard errors are barely visible, thus omitted. RankNet result for $N=10^5$ is missing as it ran out of memory.}
	\label{fig:sim-cost}
	\vskip -7pt
\end{figure*}

\begin{table*}
\centering
\caption{Precision@k and propensity-scored precision@k on the real-world XML datasets. Due to space limit, PL-Partition, RankNet, and RankSVM are respectively renamed as PL-P, R-Net, and R-SVM. \textbf{Bold numbers} indicate the best performance.}
\label{tbl:xml-p}
\scalebox{1.0}{
\begin{tabular}{llcccclcccc}
\toprule
       &      &           PL-P &  PL-LB &          R-Net & R-SVM &  &           PL-P &  PL-LB &          R-Net &          R-SVM \\
\midrule
D-1K & P@1 &  \textbf{66.72} &  66.12 &          64.11 &  61.95 &   PSP@1 &  \textbf{33.22} &  30.36 &          32.02 &          31.03 \\
       & P@3 &  \textbf{61.30} &  60.13 &          58.46 &  57.05 &   PSP@3 &  \textbf{34.23} &  31.69 &          32.76 &          31.61 \\
       & P@5 &  \textbf{56.63} &  54.72 &          53.60 &  52.74 &   PSP@5 &  \textbf{35.02} &  31.82 &          32.85 &          32.20 \\
       & P@10 &  \textbf{47.61} &  45.76 &          45.62 &  44.57 &  PSP@10 &  \textbf{35.26} &  31.91 &          33.27 &          32.71 \\
\hline
E-4K & P@1 &  \textbf{78.12} &  66.46 &          77.57 &  76.46 &   PSP@1 &          41.93 &  34.55 &  \textbf{42.81} &          42.71 \\
       & P@3 &          62.81 &  51.58 &  \textbf{63.05} &  61.88 &   PSP@3 &          43.46 &  34.37 &          45.55 &  \textbf{45.69} \\
       & P@5 &          51.75 &  41.35 &  \textbf{52.55} &  50.95 &   PSP@5 &          45.70 &  34.44 &  \textbf{48.61} &          47.43 \\
       & P@10 &          33.79 &  26.99 &  \textbf{34.36} &  32.56 &  PSP@10 &          57.13 &  42.50 &  \textbf{60.83} &          57.50 \\
\hline
W-31K & P@1 &  \textbf{85.97} &  80.73 &          82.35 &  80.88 &   PSP@1 &  \textbf{13.01} &   9.24 &          12.90 &          12.55 \\
       & P@3 &  \textbf{73.07} &  54.14 &          67.33 &  60.30 &   PSP@3 &  \textbf{13.46} &   7.61 &          12.95 &          12.18 \\
       & P@5 &  \textbf{63.01} &  44.88 &          56.96 &  50.81 &   PSP@5 &  \textbf{14.03} &   7.62 &          13.18 &          12.20 \\
       & P@10 &  \textbf{47.69} &  31.36 &          42.47 &  37.50 &  PSP@10 &  \textbf{15.70} &   7.61 &          14.90 &          13.04 \\
\hline
D-200K & P@1 &  \textbf{47.58} &  40.38 &          41.93 &  41.41 &   PSP@1 &   \textbf{8.72} &   6.79 &           7.06 &           7.13 \\
       & P@3 &  \textbf{42.09} &  37.40 &          38.92 &  38.46 &   PSP@3 &   \textbf{9.19} &   7.74 &           8.17 &           8.13 \\
       & P@5 &  \textbf{39.23} &  35.65 &          36.94 &  36.67 &   PSP@5 &   \textbf{9.82} &   8.46 &           8.93 &           8.92 \\
       & P@10 &  \textbf{35.11} &  32.40 &          33.66 &  33.44 &  PSP@10 &  \textbf{11.01} &   9.75 &          10.20 &          10.21 \\
\bottomrule
\end{tabular}
}
\end{table*}

\begin{table*}
\centering
\caption{Comparisons between the proposed PL-Paritition and SLEEC and LEML. \textbf{Bold numbers} indicate the best performance. The results of SLEEC and LEML are from the XML repository~\citep{bhatia16extreme}.}
\label{tbl:xml-sota}
\scalebox{1.0}{
\begin{tabular}{llccclccc}
\toprule
    &   & PL-Partition &  SLEEC & LEML & &  PL-Partition &  SLEEC & LEML  \\
\midrule
D-1K & P@1 &  66.72 &  \textbf{67.59} & 65.67 &   PSP@1 &  \textbf{33.22} &  32.11 &          30.73 \\
       & P@3 &  61.30 &  \textbf{61.38} &  60.55 &   PSP@3 &  \textbf{34.23} &  33.21 &          32.43 \\
       & P@5 &  \textbf{56.63} &  56.56  &  56.08 &   PSP@5 &  \textbf{35.02} &  33.83 &          33.26  \\
\hline
E-4K & P@1 &  78.12 &  \textbf{79.26} &  63.40 &   PSP@1 &  \textbf{41.93} &  34.25 &  24.10  \\
       & P@3 &          62.81 &  \textbf{64.30}  &  50.35 &   PSP@3 &          \textbf{43.46} &  39.83 &          27.20  \\
       & P@5 &          51.75 &  \textbf{52.33}  &  41.28 &   PSP@5 &          \textbf{45.70} &  42.76  &          29.09 \\
\hline
W-31K & P@1 &  \textbf{85.97} &  85.88 &  73.47 &   PSP@1 &  \textbf{13.01} &   11.14 &  9.41 \\
       & P@3 &  \textbf{73.07} & 72.98 &    62.43 &   PSP@3 &  \textbf{13.46} &   11.86 &  10.07 \\
       & P@5 &  \textbf{63.01} &  62.70  &  54.35 &   PSP@5 &  \textbf{14.03} & 12.40 & 10.55 \\
\hline
D-200K & P@1 &  47.58 &  \textbf{47.85} &   40.73 &   PSP@1 &   \textbf{8.72} &   7.17 & 6.06  \\
       & P@3 &  42.09 &  \textbf{42.21} &  37.71 &   PSP@3 &   \textbf{9.19} &   8.16 &  7.24 \\
       & P@5 &  39.23 &  \textbf{39.43}  &  35.84 &   PSP@5 &   \textbf{9.82} & 8.96 & 8.10 \\
\bottomrule
\end{tabular}
}
\vskip -5pt
\end{table*}

\vpara{Computation cost.} Figure~\ref{fig:sim-cost} shows the time and memory costs of different methods over various $N$. The results are obtained using a single Nvidia V100 GPU. We report the total time of running 1000 steps of stochastic gradient descent with batch size 20. We also report the peak CUDA memory. The costs of both time and memory for the pairwise methods grow faster than those for the listwise methods as $N$ increases.  The two listwise baseline methods, AttRank and PL-LB, have similar memory cost. PL-LB has a larger running time due to the calculation of multiple partition functions. The proposed PL-Partition method has an overhead cost due to the numerical integration. However, we observe that this overhead cost is amortized as $N$ increases. When $N=10^5$, the computational cost of PL-Partition becomes close to that of PL-LB. Overall, this benchmark empirically demonstrates that the proposed method is scalable for large-scale applications\footnote{In practice, both pairwise and listwise methods can be made more scalable by negative sampling.}.

\subsection{Real-World Datasets}
\vpara{Experiment setups.}  We also verify the effectiveness of the proposed method on 4 real-world XML datasets~\citep{bhatia16extreme}: Delicious-1K (D-1K), Eurlex-4K (E-4K), Wiki10-31K (W-31K), and Delicious-200K (D-200K). The trailing number in the name of each dataset indicates the number of classes in the dataset\footnote{The average number of labels per sample (shown in the brackets after the dataset names): D-1K (19.03), E-4K (5.31), W-31K (18.64), D-200K (75.54). See Appendix~\ref{sec:sup-xml} for more summary statistics of these datasets.}. Following many existing works~\citep{prabhu2014fastxml,bhatia2015sparse} and the official instructions of XML classification repository~\citep{bhatia16extreme}, we evaluate different methods with 4 types of ranking metrics, Precision@k, Propensity-Scored Precision@k, nDCG@k, and Propensity-Scored nDCG@k. We observe that nDCG-based metrics show similar trends compared to their Precision-based counterparts. Due to space limit, we leave the results of nDCG metrics in Appendix~\ref{sec:sup-xml}.

We first compare the proposed PL-Partition method with the 4 baseline methods. Note that PL-LB and AttRank collapse into exactly the same method on XML classification as the number of partitions is 2. So we only report the results of PL-LB. For each method, we train a neural network model with the same architecture, 2-layer fully connected network with ReLU activations and hidden size of 256. We train the neural networks with stochastic gradient descent using the ADAM optimizer. The batch size is fixed to 128. We use the official train-test split of each dataset and further split the training set into training and validation (9:1 for D-200K and 3:1 for other datasets). We tune the learning rate by line-search from $\{10^{-4}, 10^{-3}, 10^{-2}\}$ and apply early-stopping in training, based on validation sets. 

We also compare PL-Partition with two state-of-the-art embedding-based XML classifiers, \textbf{SLEEC}~\citep{bhatia2015sparse} and \textbf{LEML}~\citep{yu2014large}, which are listed on the XML repository leaderboard~\citep{bhatia16extreme}. SLEEC and LEML share similar model architectures with our setup, i.e., 2-layer neural networks, but use different training objectives: SLEEC uses a nearest-neighbor loss; LEML uses a least-square loss.

\vpara{Results.} As can be seen in Table~\ref{tbl:xml-p}, the proposed PL-Partition method significantly outperforms the softmax-based listwise method PL-LB on all datasets, indicating the importance of optimizing the proper utility function for listwise methods. PL-Partition also outperforms the pairwise methods RankSVM and RankNet on D-1K, W-31K, and D-200K, where the number of labels per sample is relatively large. When the number of labels per sample is relatively small, breaking the labels into pairwise comparisons leads to little loss of information, and pairwise methods perform well (E-4K). 

Table~\ref{tbl:xml-sota} shows the comparison between PL-Partition and embedding-based XML classifiers SLEEC and LEML. SLEEC is better than LEML on all metrics. PL-Partition achieves similar performance to SLEEC on Precision@k and significantly outperforms the baselines on Propensity-Scored Precision@k. The propensity-scored metrics are believed to be less biased towards the head items. Thus the results indicate PL-Partition has better performance than SLEEC for the torso or tail items.

\vpara{Discussions.} We note for the task of XML classification, tree-based methods~\citep{prabhu2014fastxml,jain2016extreme} sometimes outperform the embedding-based methods. The focus of this paper is to develop scalable listwise LTR methods for learning neural network ranking models from partitioned preference, instead of methods tailored for XML classification. Therefore we restrain our comparison to embedding methods only, whose model architectures are similar as the 2-layer neural networks in our experiment setup. We also note that SLEEC outperforms tree-based methods for D-200K on the XML repository leaderboard~\citep{bhatia16extreme}. This indicates PL-Partition achieves state-of-the-art performance on D-200K, where the top partition size is relatively large. \citet{guo2019breaking} recently showed that, with advanced regularization techniques, embedding-based methods trained by RankSVM or PL-LB can be significantly improved to surpass state-of-the-art tree-based methods on most XML datasets. It seems an interesting future direction to apply such regularization techniques on our proposed LTR objective.

\section{Conclusion}
In this paper, we study the problem of learning neural network ranking models with a Plackett-Luce-based listwise LTR method from data with partitioned preferences. We overcome the computational challenge of calculating the likelihood of partitioned preferences under the PL model by proposing an efficient numerical integration approach. The key insight of this approach comes from the random utility model formulation of Plackett-Luce with Gumbel distribution. Our experiments on both synthetic data and real-world data show that the proposed method is both more effective and scalable compared to popular existing LTR methods.

\subsection*{Acknowledgements}
The authors would like to thank Ao Liu, Tyler Lu, Lirong Xia, and Zhibing Zhao for helpful discussions. Jiaqi Ma and Qiaozhu Mei were in part supported by the National Science Foundation under grant numbers 1633370 and 1620319.

\bibliography{ref}
\bibliographystyle{apalike}

\clearpage
\appendix
\section{Appendix}

\subsection{Proof of Lemma~\ref{lemma:rewrite}}
\begin{proof}
To simplify the notation, let $\tilde{w}_i = \exp(w_i)$. Then
\begin{align*}
    & P(S_1\succ\cdots\succ S_M) \\
    =& \sum_{(i_1,\cdots,i_N)\in\sigma(S_1\succ\cdots\succ S_M)}\prod_{l=1}^{N}\frac{\tilde{w}_{i_l}}{\sum_{r=l}^N \tilde{w}_{i_r}} \\
    =& \sum_{\sigma(S_1)}\cdots\sum_{\sigma(S_M)}\prod_{m=1}^{M}\prod_{l=n_{m-1}+1}^{n_m}\frac{\tilde{w}_{i_l}}{\sum_{r=l}^N \tilde{w}_{i_r}}\\
    =& \prod_{m=1}^{M} \sum_{(j_1,\cdots,j_{n_m})\in\sigma(S_m)}\prod_{l=1}^{n_m}\frac{\tilde{w}_{j_l}}{\sum_{k\in R_m} \tilde{w}_k - \sum_{r=1}^{l-1}\tilde{w}_{j_r}} \\
    =& \prod_{m=1}^{M-1} P(S_m\succ R_{m+1}), 
\end{align*}
\end{proof}

\subsection{Proof of Proposition~\ref{prop:disjointab}}
\label{sec:sup-proof1}
\begin{proof}
We first show $P(A\succ B;w) = P(\min_{a\in A}g_{w_a}>\max_{b\in B}g_{w_b})$. If $A\cup B = \SN$, then the event of $A\succ B$ is equivalent to the event of $\min_{a\in A}g_{w_a}>\max_{b\in B}g_{w_b}$ so this equality holds true. Otherwise, assume there is a $c \in \SN$ but $c\notin A\cup B$. 

We introduce a few notations to assist the proof. For any $D\subseteq \SN$, let $\mathcal{G}(D) = \{g_{w_i}\mid i\in D\}$. Further let $\Omega(A\succ B; D)$ be the set of all possible permutations of $D$ that are consistent with the partial ranking $A\succ B$, i.e., 
\begin{align*}
    &\Omega(A\succ B; D) \\
    =& \{(i_1,\cdots, i_N)\in \sigma(D)\mid k < l, \forall i_k\in A, i_l\in B \}.
\end{align*}
Then we can write the LHS as
\begin{align}
    &P(A\succ B;w) \nonumber \\
    =& \sum_{\substack{(i_1,\cdots, i_N)\in \\ \Omega(A\succ B;\SN)}} P(g_{w_{i_1}} > g_{w_{i_2}} > \cdots > g_{w_{i_N}}) \nonumber \\
    =& \sum_{\substack{(i_1,\cdots, i_N)\in \\ \Omega(A\succ B;\SN)}} \idotsint_{\mathcal{G}(\SN)} \ind[g_{w_{i_1}} > g_{w_{i_2}} > \cdots > g_{w_{i_N}}] \nonumber \\
    =& \idotsint_{\mathcal{G}(\SN)} \sum_{\substack{(i_1,\cdots, i_N)\in \\ \Omega(A\succ B;\SN)}} \ind[g_{w_{i_1}} > g_{w_{i_2}} > \cdots > g_{w_{i_N}}], \nonumber
\end{align}
where we slightly abused the notation $g_{w_i}$ by using it to refer both the Gumbel random variables in the first line and the corresponding integral variables in the following lines. We have also omitted the integral variables and the probability densities $df(g_{w_i})$ in the derivation. To further ease the notation, we define $g_{w_{j_0}} = +\infty$ and $g_{w_{j_N}} = -\infty$, then
\begin{align}
    &P(A\succ B;w) \nonumber \\
    =& \idotsint_{\mathcal{G}(\SN)} \sum_{k=1}^N \sum_{\substack{(j_1,\cdots,j_{N-1})\in \\ \Omega(A\succ B;\SN\setminus \{c\})}} \nonumber \\
    &\phantomrel{=} \ind[g_{w_{j_1}} > \cdots  > g_{w_{j_{k-1}}} > g_{w_c} > g_{w_{j_{k}}} > \cdots > g_{w_{j_{N-1}}}] \nonumber \\
    =& \idotsint_{\mathcal{G}(\SN)} \sum_{\substack{(j_1,\cdots,j_{N-1})\in \\ \Omega(A\succ B;\SN\setminus \{c\})}} \ind[g_{w_{j_1}} > \cdots > g_{w_{j_{N-1}}}] \nonumber \\
    &\phantomrel{=} \cdot \sum_{k=1}^N \ind[g_{w_{j_{k-1}}} > g_{w_c} > g_{w_{j_{k}}}] \nonumber \\
    =& \idotsint_{\mathcal{G}(\SN\setminus \{c\})} \sum_{\substack{(j_1,\cdots,j_{N-1})\in \\ \Omega(A\succ B;\SN\setminus \{c\})}} \ind[g_{w_{j_1}} > \cdots > g_{w_{j_{N-1}}}] \nonumber \\
    &\phantomrel{=} \cdot \int_{g_{w_c}} \sum_{k=1}^N \ind[g_{w_{j_{k-1}}} > g_{w_c} > g_{w_{j_{k}}}], \label{eq:sum_k}
\end{align}
where the last equality utilizes the fact that all the Gumbel variables are independent.

Note that, in Eq.~(\ref{eq:sum_k}), given $g_{w_{j_1}} > \cdots > g_{w_{j_{N-1}}}$, $\sum_{k=1}^N\ind[g_{w_{j_{k-1}}} > g_{w_c} > g_{w_{j_{k}}}] \equiv 1$ regardless the choice of $(j_1,\cdots,j_{N-1})$. Therefore, $$\int_{g_{w_c}}\sum_{k=1}^N\ind[g_{w_{j_{k-1}}} > g_{w_c} > g_{w_{j_{k}}}] \equiv 1,$$ and
\begin{equation}
    \label{eq:elimination}
    \begin{split}
    &P(A\succ B;w) \\
    =& \idotsint_{\mathcal{G}(\SN\setminus \{c\})} \sum_{\substack{(j_1,\cdots,j_{N-1})\in \\ \Omega(A\succ B;\SN\setminus \{c\})}}  \ind[g_{w_{j_1}} > \cdots > g_{w_{j_{N-1}}}].
    \end{split}
\end{equation}

By applying Eq.~(\ref{eq:elimination}) to all the items that do not belong to $A\cup B$, we get
\begin{equation}
    \label{eq:marginal}
    \begin{split}
    &P(A\succ B;w) \\
    =& \idotsint_{\mathcal{G}(A\cup B)} \sum_{\substack{(j_1,\cdots,j_{|A| + |B|})\in \\ \Omega(A\succ B;A\cup B)}}  \ind[g_{w_{j_1}} > \cdots > g_{w_{j_{|A| + |B|}}}].
    \end{split}
\end{equation}
And note that this reduces to a situation equivalent to the case $A\cup B=\SN$. Therefore we have shown $P(A\succ B;w) = P(\min_{a\in A}g_{w_a} > \max_{b\in B}g_{w_b})$.

The proof for 
\[P(\min_{a\in A}g_{w_a}>\max_{b\in B}g_{w_b})=\int_{u=0}^1\prod_{a\in A}(1-u^{\exp(w_a-w_{B})}) du\]
remains the same no matter if $A\cup B=\SN$ or not, as the Gumbel variables are independent. We refer the reader to the Appendix B of~\citet{kool2020estimating} for the proof. 
\end{proof}

\subsection{Proof of Lemma~\ref{lemma:gradient}}
\begin{proof}
We first expand the gradients of the log-likelihood w.r.t. $\vw$ in Eq.~(\ref{eq:gradient}) below.
\begin{align}
\label{eq:grad}
\begin{split}
	&\nabla_{\vw} \log P (S_1 \succ  \cdots \succ S_M ;\vw) \\ 
	=& \sum_{m=1}^{M-1}\nabla_{\vw} \log P(S_m \succ R_{m+1};\vw)\\
	=&\sum_{m=1}^{M-1} \frac{1}{P(S_m \succ R_{m+1};\vw)} \nabla_{\vw} P(S_m \succ R_{m+1};\vw)\\
	=&\sum_{m=1}^{M-1} \frac{1}{P(S_m \succ R_{m+1};\vw)} \\
	& \phantomrel{=} \cdot \nabla_{\vw} \int_{u=0}^1 \prod_{i\in S_m}(1-u^{\exp(w_i-w_{R_{m+1}})}) du.
\end{split}
\end{align}

Further note that  
\begin{align}
	& \nabla_{\vw} \prod_{i\in S_m}(1-u^{\exp(w_i-w_{R_{m+1}})}) \nonumber \\
	=& \sum_{i\in S_m} \big[\prod_{j\in S_m\setminus \{i\}} (1 - u^{ \exp(w_j -w_{R_{m+1}})})\big] \nonumber \\
	& \phantomrel{=} \cdot \big[-\nabla_{\vw} u^{ \exp(w_i -w_{R_{m+1}})}\big] \nonumber \\
	=& -\sum_{i\in S_m} \big[\prod_{j\in S_m\setminus \{i\}} (1 - u^{ \exp(w_j -w_{R_{m+1}})})\big] \nonumber \\
	& \phantomrel{=} \cdot \big[ u^{ \exp(w_i -w_{R_{m+1}})} \log u \big]\nabla_{\vw}  \exp(w_i -w_{R_{m+1}}) \nonumber \\
	=& - \big[\prod_{j\in S_m} (1 - u^{ \exp(w_j -w_{R_{m+1}})})\big] \nonumber \\
	& \phantomrel{=} \cdot \sum_{i\in S_m} \frac{u^{ \exp(w_i -w_{R_{m+1}})} \log u}{1 - u^{ \exp(w_i -w_{R_{m+1}})}} \nonumber \\
	& \phantomrel{=} \cdot \nabla_{\vw}  \exp(w_i -w_{R_{m+1}}). \label{eq:grad2}
\end{align}
Plugging Eq.~(\ref{eq:grad2}) into the gradients~(\ref{eq:grad}), we have
\begin{align}
	& \nabla_{\vw} \log P (S_1 \succ  \cdots \succ S_M ;\vw) \nonumber \\
	=& - \sum_{m=1}^{M-1} \frac{1}{P(S_m \succ R_{m+1};\vw)} \nonumber \\
	& \phantomrel{=} \cdot \int_{u=0}^1 \big[\prod_{j\in S_m} (1 - u^{ \exp(w_j -w_{R_{m+1}})})\big]   \nonumber \\
	& \phantomrel{=}  \cdot \sum_{i\in S_m} \frac{u^{ \exp(w_i -w_{R_{m+1}})} \log u}{1 - u^{ \exp(w_i -w_{R_{m+1}})}} \nabla_{\vw}  \exp(w_i -w_{R_{m+1}}) du \nonumber\\
	=& - \sum_{m=1}^{M-1} \frac{1}{P(S_m \succ R_{m+1};\vw)} \sum_{i\in S_m} \nabla_{\vw}  \exp(w_i -w_{R_{m+1}}) \nonumber \\
	& \phantomrel{=} \cdot  \int_{u=0}^1 \big[\prod_{j\in S_m} (1 - u^{ \exp(w_j -w_{R_{m+1}})})\big] \nonumber \\
	& \phantomrel{=} \cdot \frac{u^{ \exp(w_i -w_{R_{m+1}})} \log u}{1 - u^{ \exp(w_i -w_{R_{m+1}})}} du. \label{eq:grad4}
\end{align}
\end{proof}

\subsection{Proof of Theorem~\ref{thm:error}}
\label{sec:thm1-proof}

Before we start our proof of Theorem~\ref{thm:error}, we first introduce the well-known discretization error bound for the composite mid-point rule of numerical integration in Lemma~\ref{lemma:cm}.
\begin{lemma} [Discretization Error Bound of the Composite Mid-point Rule.]
\label{lemma:cm}
Suppose we use the composite mid-point rule with $T$ intervals to approximate the following integral for some $x_1 > x_0$,
\[\int_{x_0}^{x_1} f(x) dx.\]
Assume $f''(x)$ is continuous for $x\in [x_0, x_1]$ and $M=\sup_{x\in [x_0, x_1]}|f''(x)|$. Then the discretization error is bounded by $\frac{M(x_1-x_0)^3}{24 T^2}$.
\end{lemma}

\vpara{Proof of the part (a).}
The sketch of the proof is as follows. We first give an upper bound of the discretization error in terms of the number of intervals. Then we can obtain the number of intervals required for any desired level of error. 

In particular, we bound the discretization error in two parts. We first bound the absolute value of the integral on the region $[0, \delta]$ for some sufficiently small $\delta>0$. We then bound the second derivative of the integrand on $[\delta, 1]$ and apply Lemma~\ref{lemma:cm} to bound the discretization error of the integral on $(\delta, 1]$. The total discretization error is then bounded by the sum of the two parts.

\begin{proof}
We first re-write the likelihood as follows,
\begin{align}
    &P (S_1 \succ  \cdots \succ S_M;\vw) \nonumber \\
    =& \prod_{m=1}^{M-1}\int_{u=0}^1\prod_{i\in S_m}\sbra1-u^{\exp\sbra w_i-w_{R_{m+1}}\sket}\sket du \nonumber \\
    =& \prod_{m=1}^{M-1} \exp(w_{R_{m+1}} + c) \nonumber \\
    & \phantomrel{=} \cdot\int_{v=0}^1 v^{\exp\sbra w_{R_{m+1}} + c\sket - 1} \prod_{i\in S_m}\sbra 1-v^{\exp\sbra w_i+c\sket}\sket dv \nonumber \\
    \triangleq & \prod_{m=1}^{M-1} I_m, \label{eq:likelihood-v-int}
\end{align}
where in the last second equality we have applied a change of variable $v = u^{\exp(-c -w_{R_{m+1}})}$ for each integral.

To simplify the notations, let us define $g_i(v) = 1-v^{\exp\sbra w_i+c\sket}$ for any $i\in S_m$, and $g_0(v) = v^{\exp\sbra w_{R_{m+1}} + c\sket - 1}$. Then $I_m$ can be written as
\[
I_m = \exp(w_{R_{m+1}} + c) \int_{v=0}^1 \prod_{i\in S_m\cup \{0\}}g_i(v) dv.
\]
Further let 
\[f(v) = \prod_{i\in S_m\cup \{0\}}g_i(v).\]
It remains to investigate the properties of $f(v)$ and its derivatives on $[0, 1]$ to bound the discretization error of $I_m$.

We first bound the absolute value of the integral on $[0, \delta]$ for some $\delta > 0$. We have
\begin{align*}
    & \left|\exp(w_{R_{m+1}} + c) \int_{v=0}^{\delta} \prod_{i\in S_m\cup \{0\}}g_i(v) dv\right| \\
    \le & C \int_{v=0}^{\delta} v^{\exp\sbra w_{R_{m+1}} + c\sket - 1} dv \\
    = & \frac{C}{\exp\sbra w_{R_{m+1}} + c\sket}\delta^{\exp\sbra w_{R_{m+1}} + c\sket}\\
    \le & \frac{C}{2}\delta^2.
\end{align*}

For any $\epsilon>0$, let $\delta=(\frac{\epsilon}{C})^{1/2}$, then \[\left|\exp(w_{R_{m+1}} + c) \int_{v=0}^{\delta} \prod_{i\in S_m\cup \{0\}}g_i(v) dv\right|\le \epsilon/2.\]

Next we bound the second derivative of the integrand, $f''(v)$, on $[\delta, 1]$. We have
\begin{align*}
    f^{''}(v) =& \sum_{i,j\in S_m\cup \{0\}, i\neq j} g'_i(v) g'_j(v) \prod_{k\in S_m\cup \{0\}\setminus \{i, j\})}g_k(v) \\
    & \phantomrel{=} + \sum_{i\in S_m\cup \{0\}} g^{''}_i(v) \prod_{k\in S_m\cup \{0\}\setminus \{i\})}g_k(v).
\end{align*}

For $v\in [\delta, 1]$, and each $i\in S_m$, we know that
\begin{align*}
    |g'_i(v)| =& \exp(w_i+c)v^{\exp(w_i + c) - 1} \\
    \le& \exp(w_i+c) \frac{1}{\delta} \\
    \le& \frac{C^{3/2}}{\epsilon^{1/2}},
\end{align*}
and 
\begin{align*}
    |g^{''}_i(v)| =& |\exp(w_i+c)^2 - \exp(w_i+c)| v^{\exp(w_i + c) - 2} \\
    \le& |\exp(w_i+c)^2 - \exp(w_i+c)| \frac{1}{\delta^2} \\
    \le& \frac{C^3}{\epsilon}.
\end{align*}
Further, 
\begin{align*}
    |g'_0(v)| =& |\exp(w_{R_{m+1}}+c) -1| v^{\exp(w_{R_{m+1}} + c) - 2} \\
    \le& \exp(w_{R_{m+1}}+c) \\
    \le& C,
\end{align*}
and 
\begin{align*}
    |g^{''}_0(v)| =& |\exp(w_{R_{m+1}}+c) -1| |\exp(w_{R_{m+1}}+c) -2| \\
	& \phantomrel{=} \cdot v^{\exp(w_{R_{m+1}} + c) - 3} \\
    \le& \exp(w_{R_{m+1}}+c)^2 \frac{1}{\delta} \\
    \le& \frac{C^{5/2}}{\epsilon^{1/2}}.
\end{align*}

Therefore, for $v\in [\delta, 1]$, we have
\begin{align*}
    |f^{''}(v)| \le& \sum_{i,j\in S_m\cup \{0\}, i\neq j} |g'_i(v)| |g'_j(v)| + \sum_{i\in S_m\cup \{0\}} |g^{''}_i(v)| \\
    \le& \frac{C^3 (n_m+1)^2}{\epsilon}.
\end{align*}

By Lemma~\ref{lemma:cm}, we know that the discretization error of the integral on $[\delta, 1]$ is bounded by
\[\frac{C^4 (n_m+1)^2}{24T^2\epsilon}.\]

For the total discretization error of $I_m$ to be smaller than $\epsilon$, it suffices to have
\[\frac{C^4 (n_m+1)^2}{24T^2\epsilon}\le \epsilon/2,\]
which means 
\[T\ge \frac{C^2(n_m+1)}{2\sqrt{3}\epsilon}.\]

\end{proof}

\vpara{Proof of the part (b).}
We follow a similar strategy as the proof of part (a). In this case, we bound the discretization error in three parts. We first bound the absolute values of the integral on the region $[0, \delta_1]$ and $[1 - \delta_2, 1]$ for some sufficiently small $\delta_1, \delta_2>0$. We then bound the second derivative of the integrand on $[\delta_1, 1-\delta_2]$ and apply Lemma~\ref{lemma:cm} to bound the discretization error of the integral on $[\delta_1, 1-\delta_2]$. The total discretization error is then bounded by the sum of the three parts.

\begin{proof}
We first denote the $(m, i)$-th integral in Eq.~(\ref{eq:grad4}) for each $m=1,\cdots, M-1$ and each $i\in S_m$ as $I_{m,i}$, i.e.,
\begin{align*}
    I_{m,i} =& \int_{u=0}^1 \big[\prod_{j\in S_m} (1 - u^{ \exp(w_j -w_{R_{m+1}})})\big] \nonumber \\
	& \phantomrel{=} \cdot \frac{u^{ \exp(w_i -w_{R_{m+1}})} \log u}{1 - u^{ \exp(w_i -w_{R_{m+1}})}} du.
\end{align*}

Similarly as what we did in the proof of part (a), by applying a change of variable $v = u^{\exp(-c -w_{R_{m+1}})}$, we can rewrite $I_{m,i}$ as
\begin{align}
    \label{eq:v}
    \begin{split}
    	I_{m,i} =& \exp(w_{R_{m+1}} + c)^2 \int_{v=0}^1 \prod_{j\in S_m} (1 - v^{\exp(w_j + c)}) \\
    	& \phantomrel{=} \cdot \frac{v^{\exp(w_i + w_{R_{m+1}} + 2c) - 1} \log v}{1 - v^{\exp(w_i + c)}} dv,
    \end{split}
\end{align}

To simplify the notation, define $g_j(v) = 1-v^{\exp\sbra w_j+c\sket}$ for any $j\in S_m$, and 
\[g_0(v) = \frac{v^{\exp(w_i + w_{R_{m+1}} + 2c) - 1} \log v}{1 - v^{\exp(w_i + c)}}.\] 
Then Eq.~(\ref{eq:v}) becomes
\[
I_{m,i} = \exp(w_{R_{m+1}} + c)^2 \int_{v=0}^1 \prod_{j\in S_m\cup \{0\}}g_j(v) dv.
\]
Further let 
\[f(v) = \prod_{j\in S_m\cup \{0\}}g_j(v).\]
Recall that we have defined $a=\exp(w_i + w_{R_{m+1}} + 2c)$ and $b = \exp(w_i + c)$ in the statement of the theorem. For ease of notation in the proof, we redefine $a=\exp(w_i + w_{R_{m+1}} + 2c) - 1$, and the assumptions (\ref{eq:condition}) become
\[a > 3, a+2b > 4, \text{ and } b> C_0.\]
Then we can simplify the notation of $g_0(v)$ as
\[g_0(v) = \frac{v^a \log v}{1 - v^b}.\]
It remains to investigate the property of $f(v)$ and its derivatives on $[0, 1]$ to bound the discretization error.

We first bound the absolute value of the integral on $[0, \delta_1]$ for some small $\delta_1>0$. We have
\begin{align*}
    & \left|\exp(w_{R_{m+1}} + c)^2 \int_{v=0}^{\delta_1} \prod_{j\in S_m\cup \{0\}}g_j(v) dv\right| \\
    \le & C^2 \int_{v=0}^{\delta_1}  \frac{v^a |\log v|}{1 - v^b} \prod_{j\in S_m}g_j(v) dv.
\end{align*}
Observe that $i\in S_m$ so $1-v^b = g_i(v)$. Further by the fact that $\log v > 1 - \frac{1}{v}$ for all $v>0$, we know that $|\log v| < |1 - \frac{1}{v}|$ for $v\in (0, 1)$. So
\begin{align*}
    & C^2 \int_{v=0}^{\delta_1}  \frac{v^a |\log v|}{1 - v^b} \prod_{j\in S_m}g_j(v) dv \\
    \le & C^2 \int_{v=0}^{\delta_1}  (v^{a-1} - v^a) \prod_{j\in S_m\setminus \{i\}}g_j(v) dv \\
    \le & C^2 \int_{v=0}^{\delta_1}  v^{a-1} dv \\
    = & \frac{C^2}{a}\delta_1^a \\
    \le & \frac{C^2}{3}\delta_1^3
\end{align*}

For any $\epsilon > 0$, let $\delta_1 = (\frac{\epsilon}{C^2})^{1/3}$, then
\[\left|\exp(w_{R_{m+1}} + c)^2 \int_{v=0}^{\delta_1} \prod_{j\in S_m\cup \{0\}}g_j(v) dv\right| \le \epsilon/3.\]

Similarly as the proof of part (a), for each $j\in S_m$ and $v\in [\delta_1, 1]$, we have 
\[|g'_j(v)|\le C/\delta_1 = C^{5/3}/\epsilon^{1/3} \le C^2/\epsilon^{1/3},\] 
and 
\[|g^{''}_j(v)|\le C^2/\delta_1^2 = C^{10/3}/\epsilon^{2/3} \le C^4/\epsilon^{2/3}.\] 
We note that, however, $g_0(v)$ is not well-defined at $v=1$. Therefore we instead try to bound the absolute value of the integral on the $[1-\delta_2, 1]$ for some small $\delta_2>0$. 

When $v$ is close to $1$, by L'Hospital's rule, we have
\begin{align*}
	\lim_{v\rightarrow 1} \frac{\log v}{1-v^b} = \lim_{v\rightarrow 1} \frac{1/v}{-bv^{b-1}} = -\frac{1}{b},
\end{align*}
and hence \[\lim_{v\rightarrow 1} g_0(v) = -\frac{1}{b}.\]

Further, from the assumptions in Eq.~(\ref{eq:cond-c}) and Eq.~(\ref{eq:condition}), we can derive that $a>b>0$. In this case, we can also show that $g'_0(v) < 0$ on $[0, 1]$ so $|g_0(v)| \le \frac{1}{b} \le \frac{1}{C_0}$.
Therefore,
\begin{align*}
	&\left|\exp(w_{R_{m+1}} +c)^2\int_{1-\delta_2}^1 f(v) dv\right|\\
	\le & C^2\int_{1-\delta_2}^1 \prod_{j\in S_m} g_j(v) |g_0(v)| dv\\
	\le &\frac{C^2}{C_0} \int_{1-\delta_2}^1 \prod_{j\in S_m} (1 - v^{\exp(w_j + c)}) dv\\
	\le &\frac{C^2}{C_0} \int_{1-\delta_2}^1 \prod_{j\in S_m} \exp(w_j + c)(1 - v) dv\\
	\le &\left[\frac{C^2}{C_0} \prod_{j\in S_m} \exp(w_j + c)\right] \delta_2^{n_m + 1}\\
	\le & \frac{C^2}{C_0} (C\delta_2)^{n_m},
\end{align*}
where for the last third inequality we have used the fact that $1-x^\alpha \le \alpha (1-x)$ when $0<x<1$ and $\alpha > 0$. 

For any $\epsilon > 0$, let 
\begin{equation}
    \label{eq:delta}
    \delta_2 = \frac{1}{C}\sbra\frac{C_0\epsilon}{3C^2} \sket^{1/n_m},    
\end{equation}
then 
\[\left|\exp(w_{R_{m+1}} +c)^2\int_{1-\delta}^1 f(v) dv\right| \le \frac{\epsilon}{3}.\]
 
Finally, we seek to bound the first and second derivatives of $g_0(v)$ on $[\delta_1, 1-\delta_2]$ in order to bound $f^{''}(v)$. We write down $g'_0(v)$ and $g^{''}_0(v)$ as follows,
\begin{align*}
    g'_0(v) = \frac{av^{a-1}\log v}{1-v^b} + \frac{v^{a-1}}{1-v^b} + \frac{bv^{a+b-1}\log v}{(1-v^b)^2},
\end{align*}
and
\begin{align*}
    g^{''}_0(v) =& \frac{(a^2-a)v^{a-2}\log v}{1-v^b} + \frac{av^{a-2}}{1-v^b} + \frac{abv^{a+b-2}\log v}{(1-v^b)^2}\\
    & \phantomrel{=} + \frac{(a-1)v^{a-2}}{1-v^b} + \frac{bv^{a+b-2}\log v}{(1-v^b)^2} \\
    & \phantomrel{=} + \frac{(b^2+ab-b)v^{a+b-2}\log v}{(1-v^b)^2} + \frac{bv^{a+b-2}}{(1-v^b)^2} \\
    & \phantomrel{=} + \frac{2ab^2v^{a+2b-3}\log v}{(1-v^b)^3}.
\end{align*}

Again we know the fact that $|\log v| < |1 - \frac{1}{v}|$ for $v\in (0, 1)$. It is also clear that $a<2C$ and $b<C$. In combination with the conditions listed in~(\ref{eq:condition}), we can bound $g_0(v)$ and its derivatives as follows,
\begin{align*}
    |g_0(v)| \le \frac{1}{1 - v^b}, |g'_0(v)| \le \frac{4C}{(1 - v^b)^2},\\
    \text{ and } |g^{''}_0(v)| \le \frac{16C^3}{(1 - v^b)^3}.
\end{align*}
We can now bound $f''(v)$ on the interval $[\delta_1, 1-\delta_2]$,
\begin{align}
    |f^{''}(v)| \le&  \sum_{i,j\in S_m\cup, i\neq j} |g'_i(v)| |g'_j(v)| |g_0(v)| \nonumber \\
    & \phantomrel{=} + \sum_{i\in S_m} \sbra |g^{''}_i(v)||g_0(v)| + |g'_i(v)| |g'_0(v)| \sket \nonumber \\
    & \phantomrel{=} + |g^{''}_0(v)| \nonumber \\
    \le& \frac{C^4 n_m^2}{(1-v^b)\epsilon^{2/3}} + \frac{4C^3 n_m}{(1-v^b)^2\epsilon^{1/3}} + \frac{16C^3}{(1-v^b)^3} \nonumber \\
    \le& \frac{16C^4n_m^2}{\epsilon^{2/3}(1-(1-\delta_2)^{C_0})^3}\nonumber \\
    \le& \frac{16C^4n_m^2}{\epsilon^{2/3}(C_0\delta_2)^3}. \label{eq:derivative-bound}
\end{align}
Plugging Eq.~(\ref{eq:delta}) into the inequality~(\ref{eq:derivative-bound}), we have
\begin{align*}
    |f^{''}(v)| \le& \frac{16C^7 n_m^2}{\epsilon^{2/3} C_0^{3+3/n_m} \frac{\epsilon}{3C^2}^{3/n_m}} \\
    \le& \frac{48C^9 n_m^2}{C_0^4 \epsilon},
\end{align*}
where for the last inequality, we have assumed $n_m \ge 9$ to ease the notation, as the case with small $n_m$ is not very interesting. 

\begin{figure*}
	\centering
	\includegraphics[width=0.32\linewidth]{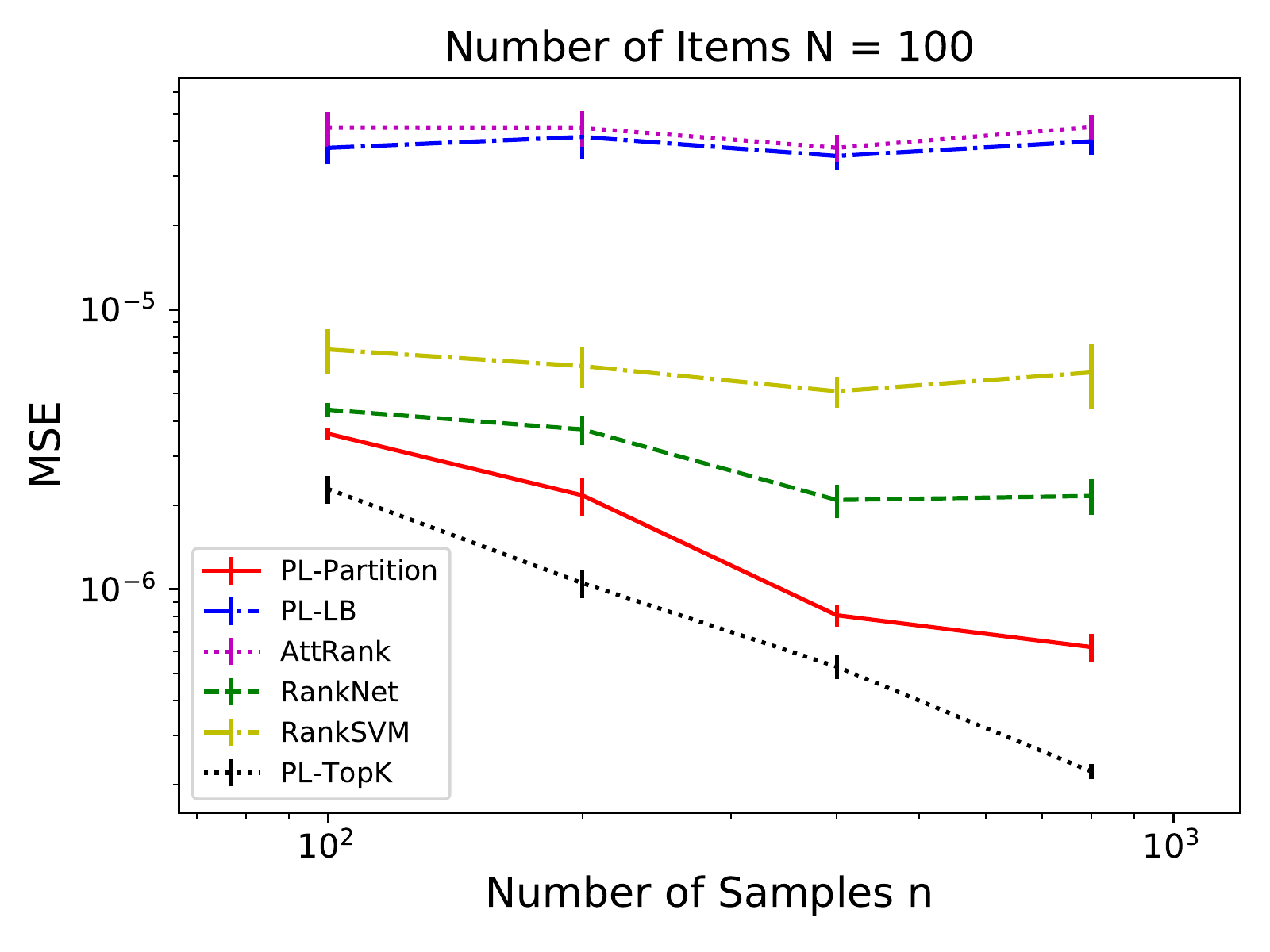}
	\includegraphics[width=0.32\linewidth]{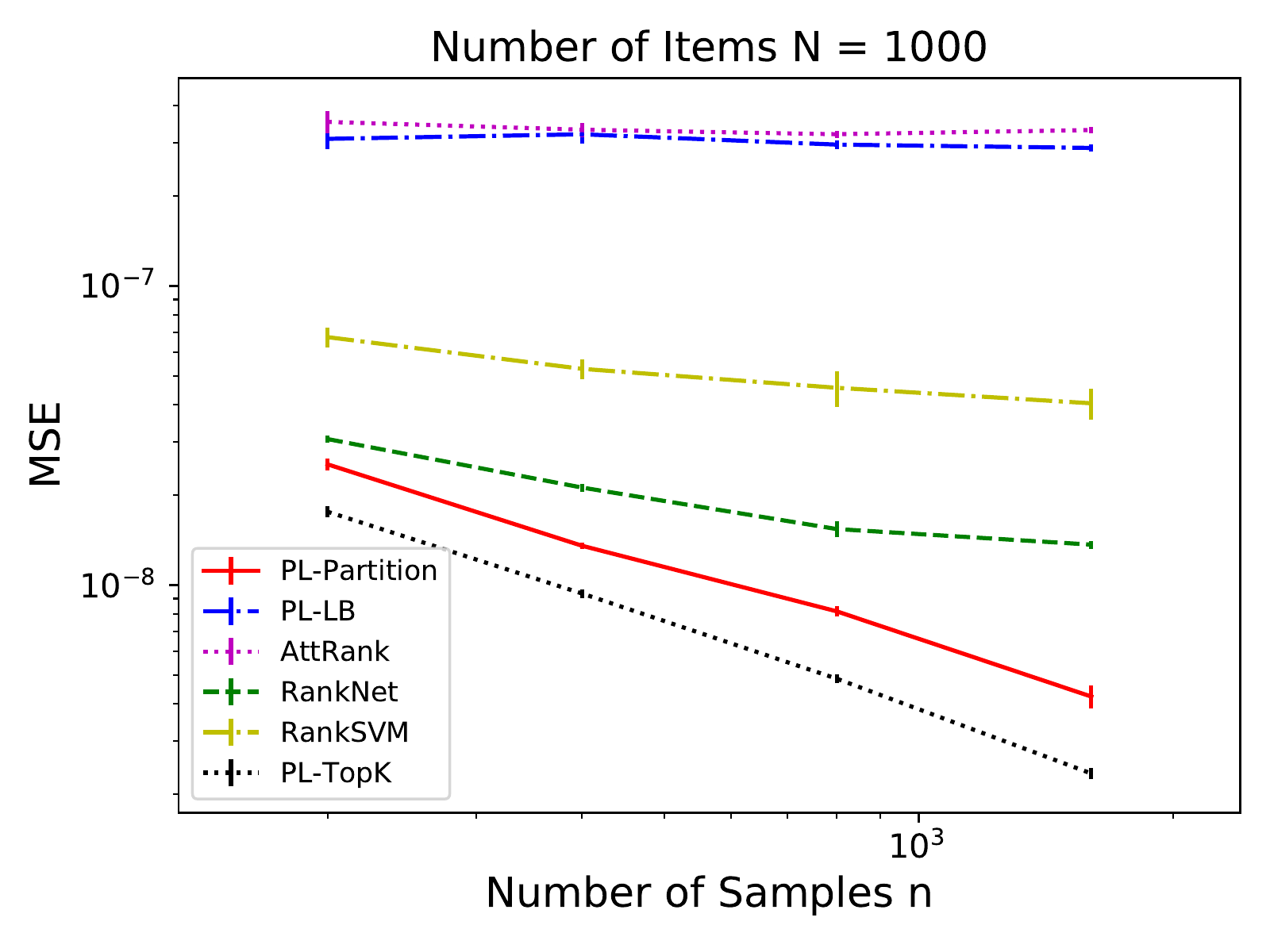}
	\includegraphics[width=0.32\linewidth]{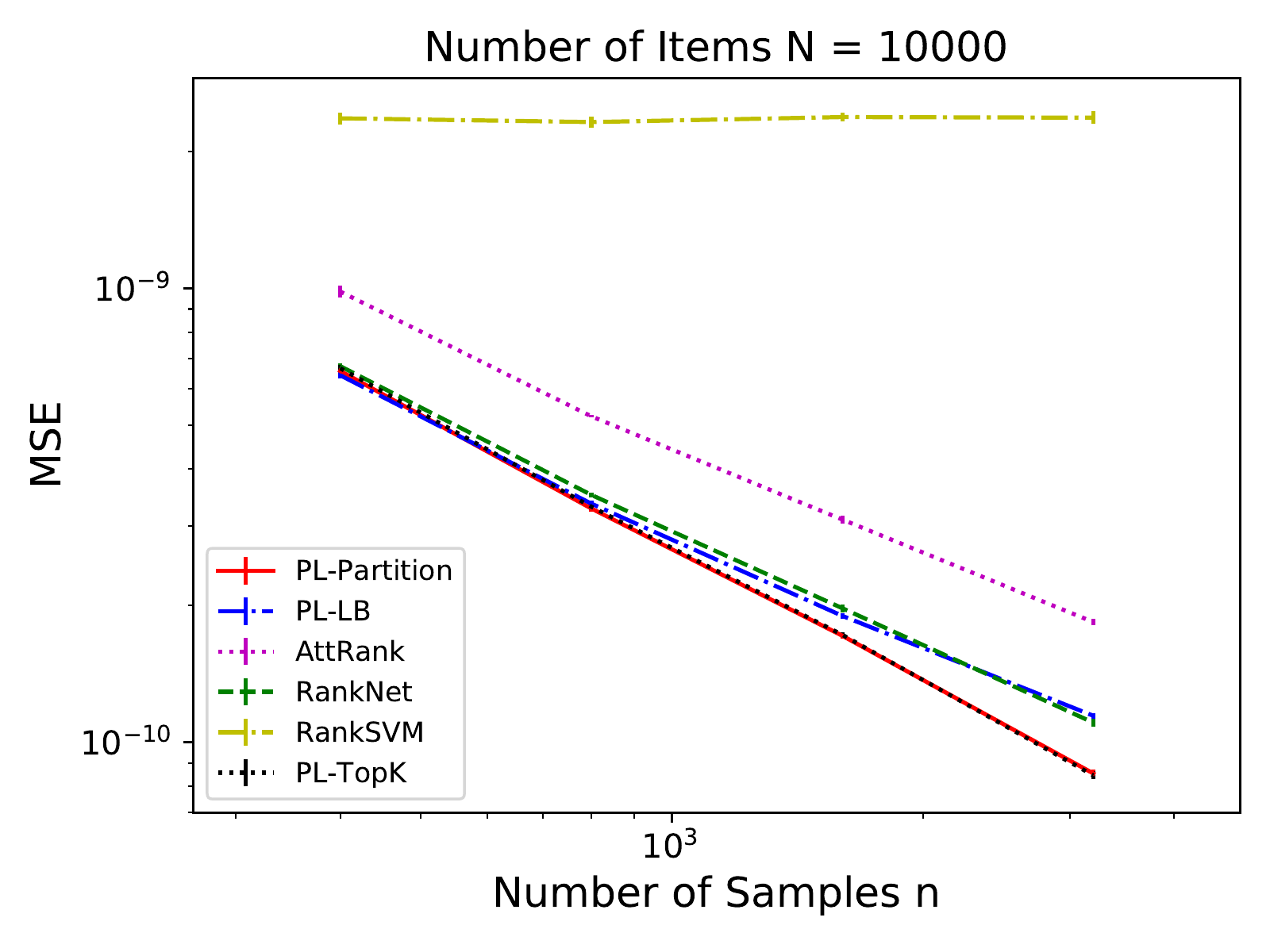}
	\caption{MSE of the estimated PL parameters vs various numbers of items $N$ and number of samples $n$. Both x-axis and y-axis are in the logarithmic scale with base 10. The results are averaged over 5 different random seeds and error bars indicate the standard error of the mean.}
	\label{fig:sup-sim-mse}
\end{figure*}

Applying the result of Lemma~\ref{lemma:cm}, we know the discretization error of the integral on $[\delta_1, 1-\delta_2]$ is bounded by
\[\frac{2C^{11} n_m^2}{T^2 C_0^4 \epsilon}.\]

For the total discretization error of $I_{m,i}$ to be smaller than $\epsilon$, it suffices to have 
\[\frac{2C^{11} n_m^2}{T^2 C_0^4 \epsilon}\le \epsilon/3,\]
which means
\[T\ge \frac{\sqrt{6} C^{11/2} n_m}{C_0^2\epsilon}.\]

To control the error of the $m$-th term in Eq.~(\ref{eq:grad4}), we may want to have discretization error of $I_{m,i}$ smaller than $\epsilon/n_m$. As an immediate corollary, we only need 
\[T\ge \frac{\sqrt{6} C^{11/2} n_m^2}{C_0^2\epsilon}.\]

\end{proof}

\subsection{Alleviate the Round-off Error}
\label{sec:sup-num}
For each integral in the likelihood~(\ref{eq:likelihood}), the integrand $\prod_{i\in S_m} (1 - u^{\exp(w_i - w_{R_{m+1}})})$ is a product of many small numbers and may suffer from round-off errors. We can alleviate such round-off errors by converting the product into summation in the logarithmic space~\citep{kool2020estimating}. In particular, recall the log-likelihood can be written as
\begin{align*}
    &\log P (S_1 \succ  \cdots \succ S_M;\vw) \\
    =&\sum_{m=1}^{M-1}\log\sbra\int_{u=0}^1\prod_{i\in S_m}\sbra 1-u^{\exp\sbra w_i-w_{R_{m+1}}\sket}\sket du\sket. 
\end{align*}
Replacing the integrals with their numerical approximation, we have
\begin{align}
    &\log P (S_1 \succ  \cdots \succ S_M;\vw) \nonumber \\
    \simeq&\sum_{m=1}^{M-1}\log\sbra\sum_{t=1}^T \prod_{i\in S_m}\sbra 1-u_t^{\exp\sbra w_i-w_{R_{m+1}}\sket}\sket \sket. \label{eq:int_sum}
\end{align}
We can rewrite Eq.~(\ref{eq:int_sum}) in the form of log-sum-exp as follows
\begin{align*}
    &\sum_{m=1}^{M-1}\log\sbra\sum_{t=1}^T \prod_{i\in S_m}\sbra 1-u_t^{\exp\sbra w_i-w_{R_{m+1}}\sket}\sket \sket \\
    =& \sum_{m=1}^{M-1}\log\sbra\sum_{t=1}^T \exp\sbra\log\sbra \prod_{i\in S_m}\sbra 1-u_t^{\exp\sbra w_i-w_{R_{m+1}}\sket}\sket \sket \sket \sket \\
    =& \sum_{m=1}^{M-1}\log\sbra\sum_{t=1}^T \exp\sbra \sum_{i\in S_m}\log\sbra 1-u_t^{\exp\sbra w_i-w_{R_{m+1}}\sket}\sket \sket \sket.
\end{align*}
Note that both the inner $\log(1-x)$ operations and the outer log-sum-exp operations have numerically stable implementations thus the influence of the round-off errors can be effectively reduced.

\subsection{Supplemental Details for Simulation}
\label{sec:sup-sim}
We also provide additional simulation results with $N=100, 1000, 10000$ in Figure~\ref{fig:sup-sim-mse}. The trend is similar as what has been shown in Figure~\ref{fig:sim-mse} in Section~\ref{sec:sim}.

\subsection{Supplemental Details for Experiments on XML Datasets}
\label{sec:sup-xml}
\begin{figure*}
    \centering
    \includegraphics[width=0.45\textwidth]{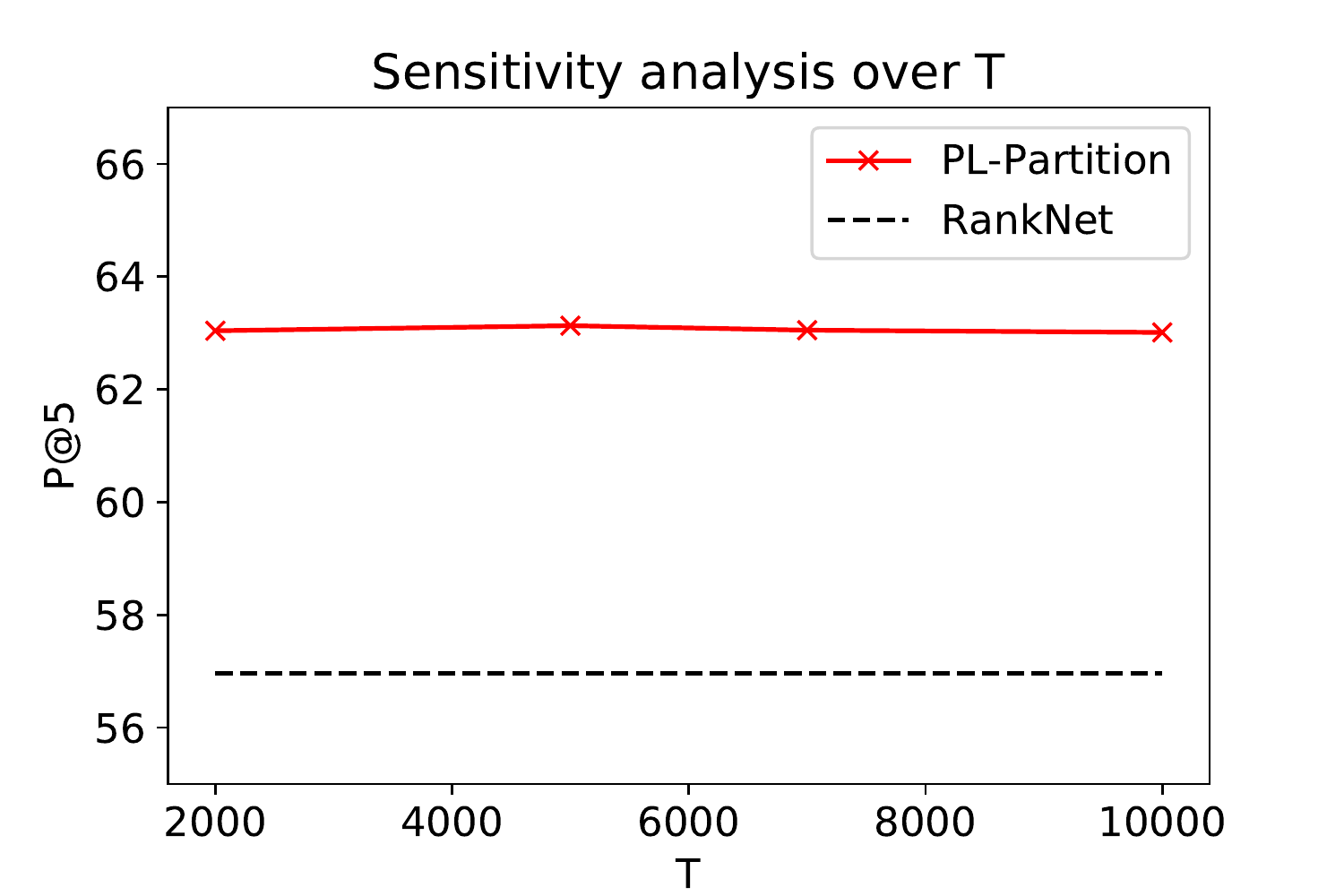}
    \includegraphics[width=0.45\textwidth]{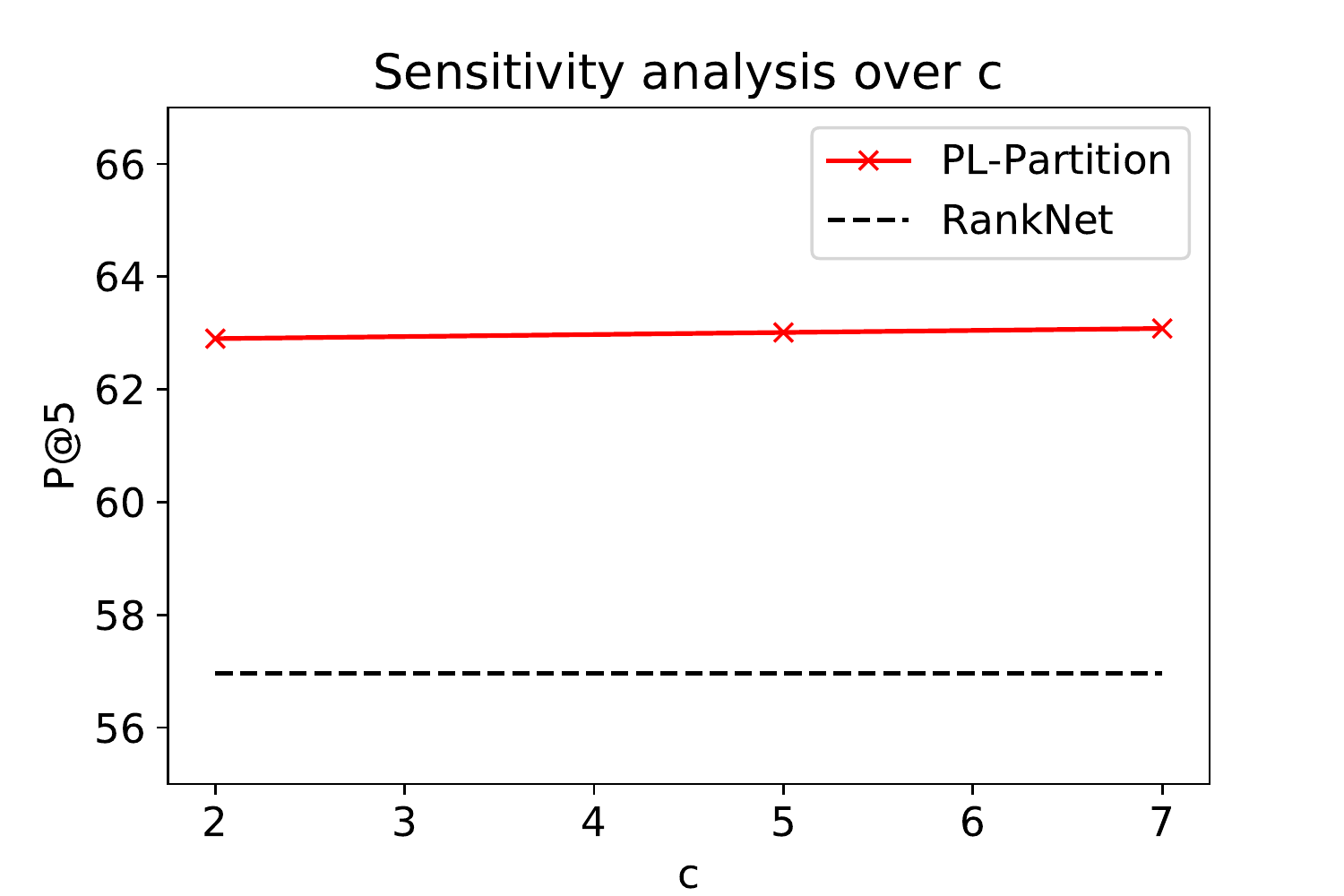}
    \caption{Sensitivity analysis over the hyper-parameters $T$ and $c$ of the proposed PL-Partition on the W-31K dataset. In both plots, the x-axis is the hyper-parameter value, and the y-axis is the P@5 metric. As a reference, we also plot the P@5 of the baseline RankNet, which is the black dashed constant line.}
    \label{fig:sensitivity}
\end{figure*}

We provide the summary statistics of the 4 XML classification datasets in Table~\ref{tab:sup-stats-dataset}. We also provide the results of the nDCG-based metrics in Table~\ref{tbl:xml-ndcg}. The nDCG-based metrics are highly correlated with their Precision-based counterparts. We further display in Figure~\ref{fig:sensitivity} that the proposed PL-Partition is not sensitive in a wide range of the hyper-parameters $T$ and $c$.

\begin{table*}
    \centering
    \caption{Summary Statistics of the XML Classification Datasets}
    \label{tab:sup-stats-dataset}
    \scalebox{0.9}{
    \begin{tabular}{lcccccc}
        \toprule
        Dataset & \#Feature & \#Label & \#Train & \#Test & Avg. \#Sample per Label & Avg. \#Label per Sample \\
        \midrule
        D-1K & 500 & 983 & 12920 & 3185 & 311.61 &  19.03 \\
        \hline
        E-4K & 5000 & 3993 & 15539 & 3809 & 25.73 & 5.31 \\
        \hline
        W-31K & 101938 & 30938 & 14146 & 6616 & 8.52 & 18.64 \\
        \hline
        D-200K & 782585 & 205443 & 196606 & 100095 & 72.29 & 75.54 \\
        \bottomrule
    \end{tabular}
    }
\end{table*}

\begin{table*}
\centering
\caption{nDCG@k and propensity-scored nDCG@k on the real-world XML datasets. Due to space limit, PL-Partition, RankNet, and RankSVM are respectively renamed as PL-P, R-Net, and R-SVM. \textbf{Bold numbers} indicate the best performance.}
\label{tbl:xml-ndcg}
\scalebox{0.8}{
\begin{tabular}{llcccclcccc}
\toprule
       &         &           PL-P &  PL-LB &          R-Net & R-SVM &  &           PL-P &  PL-LB &          R-Net &          R-SVM \\
\midrule
D-1K & nDCG@1 &  \textbf{66.72} &  66.12 &          64.11 &  61.95 &   PSnDCG@1 &  \textbf{33.22} &  30.36 &          32.02 &          31.03 \\
       & nDCG@3 &  \textbf{62.78} &  61.57 &          60.00 &  58.24 &   PSnDCG@3 &  \textbf{36.01} &  33.26 &          34.34 &          33.29 \\
       & nDCG@5 &  \textbf{59.25} &  57.51 &          56.29 &  55.01 &   PSnDCG@5 &  \textbf{37.58} &  34.55 &          35.77 &          34.87 \\
       & nDCG@10 &  \textbf{52.48} &  50.70 &          50.23 &  49.08 &  PSnDCG@10 &  \textbf{39.48} &  36.09 &          37.43 &          36.52 \\
\hline
E-4K & nDCG@1 &  \textbf{78.12} &  66.46 &          77.57 &  76.46 &   PSnDCG@1 &          41.93 &  34.55 &  \textbf{42.81} &          42.71 \\
       & nDCG@3 &          66.40 &  55.16 &  \textbf{66.56} &  65.32 &   PSnDCG@3 &          47.61 &  38.18 &          49.83 &  \textbf{49.87} \\
       & nDCG@5 &          59.66 &  48.71 &  \textbf{59.86} &  58.61 &   PSnDCG@5 &          50.51 &  39.19 &  \textbf{52.98} &          52.50 \\
       & nDCG@10 &          57.34 &  46.78 &  \textbf{57.78} &  56.35 &  PSnDCG@10 &          50.76 &  39.19 &  \textbf{53.33} &          52.72 \\
\hline
W-31K & nDCG@1 &  \textbf{85.97} &  80.73 &          82.35 &  80.88 &   PSnDCG@1 &  \textbf{13.01} &   9.24 &          12.90 &          12.55 \\
       & nDCG@3 &  \textbf{76.08} &  59.79 &          70.88 &  64.39 &   PSnDCG@3 &  \textbf{14.41} &   8.72 &          14.07 &          13.32 \\
       & nDCG@5 &  \textbf{68.40} &  51.85 &          62.81 &  56.67 &   PSnDCG@5 &  \textbf{15.98} &   9.25 &          15.31 &          14.33 \\
       & nDCG@10 &  \textbf{56.17} &  40.24 &          50.71 &  45.56 &  PSnDCG@10 &  \textbf{18.98} &  10.22 &          17.99 &          16.46 \\
\hline
D-200K & nDCG@1 &  \textbf{47.58} &  40.38 &          41.93 &  41.41 &   PSnDCG@1 &   \textbf{8.72} &   6.79 &           7.06 &           7.13 \\
       & nDCG@3 &  \textbf{43.32} &  37.85 &          39.68 &  39.24 &   PSnDCG@3 &   \textbf{9.98} &   8.27 &           8.67 &           8.70 \\
       & nDCG@5 &  \textbf{41.07} &  36.50 &          38.09 &  37.77 &   PSnDCG@5 &  \textbf{10.98} &   9.27 &           9.79 &           9.79 \\
       & nDCG@10 &  \textbf{37.69} &  33.99 &          35.55 &  35.20 &  PSnDCG@10 &  \textbf{12.51} &  10.83 &          11.39 &          11.42 \\
\bottomrule
\end{tabular}
}
\end{table*}

\end{document}